\documentclass[a4paper]{article}
\usepackage[a4paper,top=3cm,left=3cm,right=2cm,bottom=2cm]{geometry}
\usepackage[affil-it]{authblk}

\usepackage{amsthm}
\usepackage{natbib}
\usepackage{fancyhdr}
\usepackage{colortbl}
\usepackage{amsmath}
\usepackage{multirow}
\usepackage[noend]{algpseudocode}
\usepackage[usenames,dvipsnames]{xcolor}
\usepackage[vertfit]{breakurl} 
\usepackage{booktabs}
\usepackage{tikz}
\usepackage{framed}
\usepackage{adjustbox}
\usetikzlibrary{arrows}

\usepackage{lscape}
\usepackage{graphics}
\usepackage{graphicx}
\usepackage{caption}
\usepackage{subcaption}
\usepackage{csquotes}
\usepackage{longtable}
\usepackage{url}
\usepackage[shortlabels]{enumitem}
\usepackage{bbm, calrsfs, mathrsfs}
\usepackage{array}
\usepackage{rotating}
\usepackage[normalem]{ulem}
\usepackage[titletoc,title]{appendix}
\usepackage{bigstrut}
\usepackage[linesnumbered,ruled,vlined]{algorithm2e}
\usepackage{setspace}

\newcolumntype{L}[1]{>{\raggedright\let\newline\\\arraybackslash\hspace{0pt}}m{#1}}
\newcolumntype{C}[1]{>{\centering\let\newline\\\arraybackslash\hspace{0pt}}m{#1}}
\newcolumntype{R}[1]{>{\raggedleft\let\newline\\\arraybackslash\hspace{0pt}}m{#1}}

\newcommand{\GG}[1]{}

\DeclareMathOperator*{\argmin}{argmin} 

\newcommand{\up}[1]{\raisebox{1.3ex}[0pt]{#1}}



\begin{document}
\sloppy
\allowdisplaybreaks
\newcolumntype{H}{>{\setbox0=\hbox\bgroup}c<{\egroup}@{}}




\large
\title{Exact and heuristic methods for the discrete parallel machine scheduling location problem}

\author[1]{Raphael Kramer\thanks{ email: \texttt{raphael.kramer@ufpe.br}}}
\author[2]{Arthur Kramer\thanks{email: \texttt{arthur.kramer@ct.ufrn.br}}}

\affil[1]{{\small Departamento de Engenharia de Produ\c{c}\~{a}o\\ Universidade Federal de Pernambuco, Brazil}}
\affil[2]{{\small Departamento de Engenharia de Produ\c{c}\~{a}o\\ Universidade Federal do Rio Grande do Norte, Brazil}}

\date{}

\maketitle

\vspace{-0.75cm}
\begin{center}
 Technical report -- June 2020 \\
\end{center}

\begin{abstract}
	The discrete parallel machine makespan scheduling location (ScheLoc) problem is an integrated combinatorial optimization problem that combines facility location and job scheduling. The problem consists in choosing the locations of $p$ machines among a finite set of candidates and scheduling a set of jobs on these machines, aiming to minimize the makespan. Depending on the machine location, the jobs may have different release dates, and thus the location decisions have a direct impact on the scheduling decisions. To solve the problem, it is proposed a new arc-flow formulation, a column generation and three heuristic procedures that are evaluated through extensive computational experiments. By embedding the proposed procedures into a framework algorithm, we are able to find proven optimal solutions for all benchmark instances from the related literature and to obtain small percentage gaps for a new set of challenging instances.
\end{abstract}

%
\onehalfspace
\section{Introduction}
\label{sec:intro} 

Scheduling and facility location represent two classes of well-studied combinatorial optimization problems. The main motivation for studying them relies on the broad range of applications (e.g., in public services, industry, logistics, project management, production planning, data processing, etc.), as well as on the challenge in providing efficient solutions, since many of these problems are classified as NP-hard (see, e.g., \citealt{Pinedo2009}, \citealt{Pinedo2016}, \citealt{DreznerHamacher2002}, and \citealt{Laporte2015a}). Since the 1960s, many works on these topics have been published, but only a few of them has focused on studying these problems in an integrated fashion. 
Due to the limited capacity of the computers of two decades ago, it was usual to solve integrated combinatorial optimization problems using sequential approaches, i.e., solving each problem separately in such a way that the solution of one represents an input to the other. 
However, this strategy does not guarantee the optimality of the overall solution and, in addition, the input solutions may not be feasible for the successor problems. 
With the recent advances in technology, especially in the computational field, solving integrated combinatorial optimization problems using integrated approaches is becoming more accessible.

In this context, the ScheLoc problem combines the job scheduling and facility location in a single and integrated problem. It consists in selecting the locations of $p$ machines, and in scheduling a set of jobs to be processed on these machines. In order to process a job, first, it must be transported (e.g., from a warehouse or an external supplier) to the location of the machine that will process it. The time required for this transportation defines the earliest time to start processing the job in that machine. 
A straightforward application of this problem is the use of movable machines in production planning, whose locations may vary according to the scheduling objective \citep{HennesHamacher2002}.
Other applications include the location of crushers for processing minerals, the selection of the best position to dock a ship to be loaded, and the location of training devices for military forces \citep{Elvikis2009, Kalsch2009, Kalsch2010, Hessler2017}.

In this work, we propose exact and heuristic optimization methods to obtain good lower bounds and upper bounds to the \emph{discrete parallel machine makespan} (DPMM) ScheLoc problem. The main contributions of this paper are listed in the following:

\begin{itemize}
	\itemsep0em
	\item We propose a new \emph{arc-flow} (AF) formulation to model and solve the integrated DPMM ScheLoc problem, by using a pseudo-polynomial number of variables and constraints.
	\item A simple and effective column generation procedure is designed to solve the linear relaxation of the proposed AF formulation. Due to the pseudo-polynomial size of the formulation, this method makes it possible to obtain good lower bounds for large scale instances.
	\item Two integer programming-based heuristics that rely on the solution of restricted and smaller AF formulations are developed to obtain good upper bounds.
	\item A third heuristic method, based on the iterated local search metaheuristic, is also developed. This method makes use of auxiliary data structures to speed-up the local search and to provide high quality feasible solutions in short computational times.
	\item The mentioned methods are embedded into a framework algorithm developed to extract their strengths and solve the DPMM ScheLoc problem.
	\item Extensive computational experiments are carried out to assess the performance of our proposed algorithms, demonstrating that they outperform the state-of-the-art methods from the literature. In particular, the framework was capable of proving the optimality of all existing benchmark instances proposed by \citet{Hessler2017}.
	\item A new set of challenging instances are proposed.
\end{itemize}

The remainder of this paper is organized as follows. Section \ref{sec:litreview} provides a literature review on related works. Section \ref{sec:formulations} formally describes the DPMM ScheLoc problem and presents our proposed AF formulation, as well as a column generation algorithm employed to solve the linear relaxation of the proposed model. 
In Section \ref{sec:heuristics}, we detail the proposed heuristic methods, i.e., the two integer programming-based heuristics and the iterated local search metaheuristic.
The complete framework algorithm for solving the problem is presented in Section \ref{sec:exactframework}.
Computational experiments and the results obtained by our methods are presented and discussed in Section \ref{sec:results}. Finally, Section \ref{sec:conclusions} concludes the paper and indicates further research directions.

\section{Literature review}
\label{sec:litreview}

The ScheLoc problem was formally introduced by \citet{HennesHamacher2002} with the aim of minimizing the makespan. In this first version, the jobs must be scheduled in a single machine, and the machine can be located anywhere on a network. \citet{Kaufmann2014} also studied the single machine ScheLoc problem with candidate locations restricted to a network. Still, they considered a universal objective function that contains as special cases the minimization of makespan and the minimization of the total completion time. The single machine ScheLoc makespan problem in which machines can be located in any position of a plane was investigated by \citet{Elvikis2009}. They proposed two global search procedures and one local search heuristic for solving it. \citet{Kalsch2010} studied the single machine ScheLoc problem considering two objective functions (the makespan minimization and the total completion time minimization) and proposed a branch-and-bound approach for solving instances containing up to 10,000 jobs with Euclidean, rectilinear and general $\ell_q $ distances. It is worth mentioning that, once the machine location is defined, the single machine makespan ScheLoc problem can be solved to optimality in polynomial time by means of the \emph{earliest release date} (ERD) rule, i.e., by sequencing the jobs in non-decreasing order of release dates \citep{Lawler1973, Brucker2007, Pinedo2016}. 

\citet{Hessler2017} proposed the \emph{discrete parallel machine makespan} (DPMM) ScheLoc problem that differs from the previous versions by considering $p$ parallel machines whose locations should be selected from a discrete set of candidate locations. In addition to the machine location decisions, the DPMM ScheLoc problem involves the assignment of jobs to locations (or machines) and the scheduling of jobs over the $p$ machines. To solve the problem, they proposed a mathematical formulation and some heuristics. Concerning the heuristics, they first generate initial solutions using different clustering procedures. Then, from the obtained solutions, a post-procedure is applied to mitigate the load unbalance of the machines. Also, a local search algorithm based on the ERD rule and some lower bounds are presented. 

More recently, \citet{Wang2020} presented a network flow-based formulation and three heuristic procedures to solve the DPMM ScheLoc problem. Based on computational experiments, the authors show that, within a time limit of one hour, the proposed formulation is able to solve more instances to optimality than the formulation presented by \citet{Hessler2017}. However, the percentage gaps for the instances involving more than $20$ jobs are remarkably high ($>40\%$). 
Regarding the heuristic solutions, the results are also better than those reported by \citet{Hessler2017}, but the percentage gaps to the best known lower bounds are still high (around $9\%$ on average).

\citet{Kerkhove2014a} presented a case study at a Belgian textile manufacturer whose production lines are situated in dispersed locations, each of which containing multiple machines operating in parallel. The studied problem has several practical constraints, which include machine-dependent release dates and due dates. However, the location of the machines are not decisions to be taken. Moreover, after processing a job $j$ in a machine $k$, the job must be transported from the machine location to a delivery location, satisfying a due date. If the due date is violated, a penalty cost must be paid. Additional features of the problem include sequence-dependent setup times and machine-dependent processing times (i.e., unrelated machines). The authors propose a hybrid metaheuristic that combines characteristics of simulated annealing and genetic algorithm to solve instances involving up to $750$ jobs, $75$ machines, and $10$ production locations. Later, \citet{Kerkhove2014b} showed through computational experiments that using combined neighborhood structures usually leads to better results than using single neighborhood structures.

The DPMM ScheLoc problem with unrelated machines has been presented by \citet{Lawrynowicz2019}. They proposed a tabu search based memetic algorithm to solve the integrated problem and showed that it performs better than solving the location and scheduling problems sequentially. The gamma heuristic proposed by \cite{Rosing1999} is used to define the machine locations, and the particle swarm optimization metaheuristic proposed by \cite{Lin2013} is used to schedule the jobs. They also solved the DPMM ScheLoc problem with identical machines, and the obtained solutions outperformed those generated by the clustering heuristic of \citet{Hessler2017}. 

Considering uncertainty on job-processing times, \citet{Liu2019} studied a two-stage stochastic version of the discrete parallel machine ScheLoc problem aiming at minimizing the weighted sum of the location cost and the expected total completion time. In the first stage, the decisions related to machine locations are taken, while in the second stage, when the full information of processing times is obtained, the scheduling problem is solved. \citet{Krumke2020} also investigate the ScheLoc with uncertain job-processing times, but involving only a single machine and aiming at minimizing the makespan value in the worst-case.

\section{Problem description and mathematical formulations}
\label{sec:formulations}

\begin{figure}[htbp]
	\centering
	\includegraphics[scale=0.255]{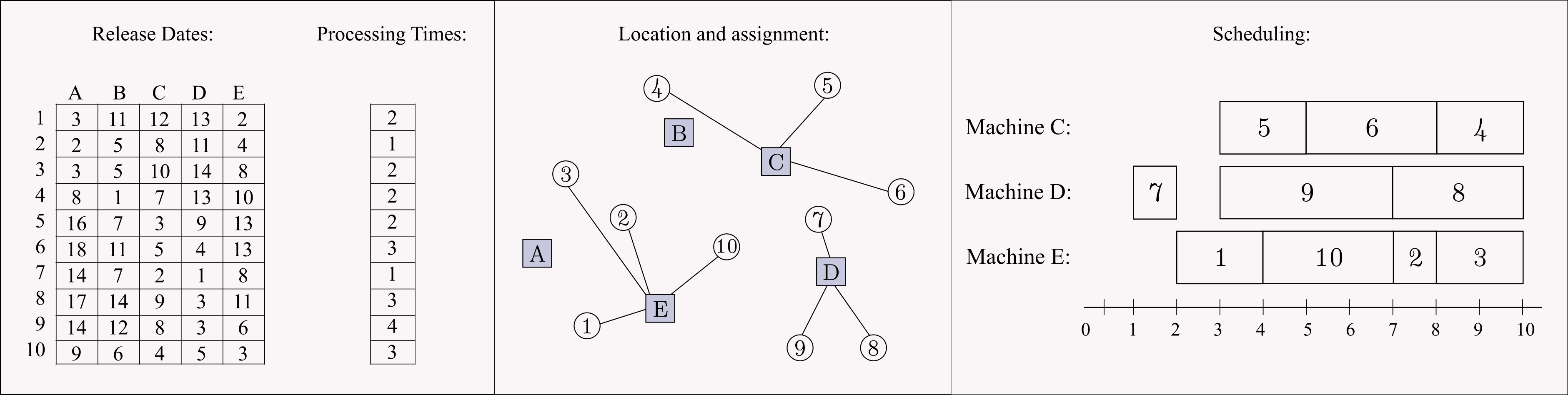}
	\caption{Solution representation for an instance with $n=10$, $m=5$, and $p=3$.}
	\label{fig:example}
\end{figure}

In the DPMM ScheLoc problem, we are given a set $J = \{1,\dots,n\}$ of jobs to be scheduled in at most $p$ identical machines whose locations must be chosen from a discrete set $M = \{1,\dots,m\}$ of candidates (with $p < m$). Each job $j \in J$ must be processed for $p_j$ units of time by one of the $p$ machines, without preemption, and only one machine can be located in a candidate location $k \in M$. 
Moreover, a job $j$ can only be processed in a location $k$ from a given release date $r_{jk}$, which can be associated with the transportation time from the job storage location to the machine location. The objective consists in minimizing the makespan, i.e., the completion time of the last processed job. The DPMM ScheLoc problem is NP-hard since it generalizes the parallel machine makespan scheduling problem, which is known to be NP-hard (\citealt{GareyJohnson1990}).
An illustrative example of a toy instance and its optimal solution is depicted in Figure \ref{fig:example}. The left part of the figure shows the instance data, while the central and right parts show the optimal locations, assignments, and scheduling. 
Mathematical formulations for the DPMM ScheLoc problem have already been proposed by \citet{Hessler2017} and \citet{Wang2020}. In addition to these, we present next a new arc-flow (AF) formulation whose linear relaxation provides good lower bounds.

\subsection{Arc-flow formulation}
\label{sec:arcflow}

AF formulations have been successfully applied to a variety of combinatorial optimization problems, such as bin packing and cutting stock (see e.g., \citealt{ValeriodeCarvalho1999, Delorme2016}), berth allocation (see e.g., \citealt{KramerLIV_2019}), vehicle routing (see e.g., \citealt{Macedoetal2011}), facility location (see e.g., \citealt{Kramer2020}), and scheduling problems (see e.g., \citealt{MradAndSouayah2018, KramerDI_2019, KramerIL_2019}).
These formulations are known for modeling combinatorial optimization problems by using flows on a capacitated network composed of a source and a sink node as well as of intermediate nodes, for each available resource. The flows from the source to the sink node are decomposed into paths that represent the solutions. In scheduling problems, the resources are the machines, the nodes represent time instants, the arcs represent the processing of jobs at a specific time, and a path from the source to the sink node represents a machine schedule.
AF formulations make use of a pseudo-polynomial number of variables and constraints such as in the time-indexed formulations (see, e.g., \citealt{SousaWolsey1992}). Indeed, the equivalence of both models are shown in \cite{VdC2002}, \cite{KramerDI_2019} and \cite{KramerLIV_2019}. Due to the pseudo-polynomial size of AF formulations, a remarkable effort is commonly expended to reduce the number of variables and constraints.

Our AF formulation models the DPMM ScheLoc problem as the problem of finding at most $p$ independent paths from the source node $0$ to the sink node $T$, containing at least one arc associated to each job $j \in J$, where $T$ represents the end of the time horizon, i.e., an upper bound to $C_{\max}$.
The proposed AF formulation makes use of a direct acyclic multigraph $G=(N,A)$. The set of nodes is defined by $N = \cup_{k \in M}N_k$, where $N_k$ represents the possible start and completion times for the jobs on machine $k \in M$, e.g., $N_k = \{0,1,\dots,T\}$. The set of arcs $A$ represent the many possibilities for processing the jobs on the machines.
It can be partitioned as $A = \cup_{k \in M} A_k$, with $A_k = \cup_{j \in J\cup \{0\}}A_{jk}$, where $A_{jk}$ refers to the set of arcs associated to a job $j \in J$ and a machine $k$, and $A_{0k}$ represents the set of dummy arcs, used to model idle times.
Formally, $A_{jk} = \{(q,r,j,k): q \in N_k, r_{jk}\leq q \leq T - p_j, \: r = q + p_j\}$, where $(q,r,j,k)$ denotes an arc associated to the processing of job $j$ on machine $k$, starting at time (node) $q$ and ending at time (node) $r=q+p_j$. Regarding the set of dummy arcs, let us first define the set $R_k = \{r_{jk}, j \in J\}\cup \{T\}$ as the set containing the distinct release dates of jobs $j \in J$ for machine $k \in M$ including the time $T$. Then, $A_{0k} = \{(q,r,0,k): q \in N_k \setminus\{T\}, \text{ and } r \text{ is the smallest value in } R_k \text{ greater than } q\}$.	

Usually, AF formulations make use of a subset $N' \subseteq N$ that relies on the so-called normal patterns (see, e.g., \citealt{CoteIori2018}), with $N' = \cup_{k \in M}N'_k$, where $N'_k \subseteq N_k$. In our AF model, in addition to the consideration of the normal patterns, we make use of an ordering of jobs to further reduce the size of set $N'$ that is based on the \emph{earliest release date} (ERD) rule. This procedure works similarly to the one adopted by \citet{KramerDI_2019} in their first reduction of variables and constraints. As mentioned before, the ERD rule is known for being able to solve the $1|r_j|C_{\max}$ problem to optimality (see, e.g., \citealt{Lawler1973, Pinedo2016, Brucker2007}). Concerning the DPMM ScheLoc problem, it is known that there exists at least one optimal solution in which the sequence of jobs on each machine follows the ERD rule (see, e.g., \citealt{Elvikis2009}). In summary, the procedure adopted to obtain the graph $G=(N',A)$ is presented in Algorithm \ref{algo:normal_patterns}.

\begin{algorithm}
	\footnotesize
	{\bf Initialize $P[1,\dots,m][0,\dots,T] \gets \texttt{false}$;}\\
	{\bf Initialize $A[1,\dots,m][0,\dots,n] \gets \emptyset$;}\\
	{\bf Initialize $N'[1,\dots,m] \gets \{T\} $;}\\
	\For{$k \in M$}
	{
		Sort $J$ according to the ERD rule for machine $k$;\\
		\For{$j \in J$}
		{
			{$P[k][r_{jk}] \gets \texttt{true}$;}\\
			\For{$t = T-p_j$ down to $r_{jk}$}
			{
				\If{$P[k][t] = \text{true} $}
				{
					$P[k][t+p_j] \gets \text{true}$;\\ 
					$A[k][j] \gets A[k][j] \cup \{(t,t+p_j,j,k)\}$;
				}
			}
		}
		\textbf{Initialize} $R$ as a vector containing all elements from $R_k$ sorted in increasing order\\
		$i \gets 1$;\\
		\For{$t = 0$ to $T-1$}
		{
			\If{$t \geq R[i]$}{$i \gets i + 1$}
			\If{$P[k][t] = \texttt{true}$}{
				$N'[k] \gets N'[k] \cup \{t\}$;\\
				$A[k][0] \gets A[k][0] \cup \{(t,R[i],0,k)\}$;
			}
		}
	}
	{\bf return} $N',A$;
	\caption{{\sc Create AF multigraph}}
	\label{algo:normal_patterns}
\end{algorithm}

In order to model the DPMM ScheLoc problem by means of an AF formulation, we associate a variable $x_{qrjk}$ with each arc $(q,r,j,k) \in A$. This variable assumes value $1$ if a job $j \in J$ is processed by a machine in location $k$ from time $q$ to time $r$, $0$ otherwise; while for $j = 0$ (i.e., for the dummy arcs) this variable is continuous and can assume any value in the interval $[0,1]$. We also introduce the binary variables $y_k, k \in M$, which take value $1$ if a machine is located in $k$, $0$ otherwise, and the continuous variable $C_{\max}$ to represent the makespan. Considering these definitions, the AF formulation for the DPMM ScheLoc problem is as follows:
\vspace{-0.25cm}
\begin{align}
(\mbox{AF}) \quad \min \quad C_{\max} \label{FO:AF}
\end{align}
\vspace{-0.3cm}
s.t.
\vspace{-0.5cm}
\begin{align}
\sum_{(q,r,j,k) \in A} x_{qrjk} \geq 1 & & &  j \in J,
\label{constr1:AF}\\
\sum_{(r,s,j,k) \in A } \hspace{-0.1cm} x_{rsjk} - \hspace{-0.3cm} \sum_{(q,r,j,k) \in A} \hspace{-0.1cm} x_{qrjk} = \left\{
\begin{array}{l l}
y_k, & \text{ if $r = \min_{j \in J}\{r_{jk}\}$}\\
-y_k, & \text{ if $r = T$}\\
0, & \text{otherwise}
\end{array}\right. &&& k \in M, r \in N'_k,\label{constr2:AF}\\
\sum_{(q,r,j,k) \in A} r \: x_{qrjk} \leq C_{\max} & & & j \in J,\label{constr3:AF}\\
\sum_{k \in M} y_k \leq p, \label{constr4:AF}\\
C_{\max} \leq T, \label{constr5:AF}\\
y_k \in \{0,1\} &&& k \in M, \label{constr7:AF}\\
0 \leq x_{qr0k} \leq 1 &&& (q,r,0,k) \in A, \label{constr6:AF}\\
x_{qrjk} \in \{0,1\} &&& j \in J, (q,r,j,k) \in A.\label{constr8:AF}
\end{align}

\begin{figure}[!ht]
	\centering
   	\begin{tikzpicture}[thick, scale=0.65, every node/.style={transform shape}, >=stealth', dot/.style = {draw, fill = white, circle, inner sep = 0pt, minimum size = 4pt}]
	\draw[very thin, color=black!50] (0,0) -- (24,0);\draw[very thin, color=black!50] (0,-0.35) -- (0,0.35);\draw[very thin, color=black!50] (24,-0.35) -- (24,0.35);
	\draw[very thin, color=black!50] (0,4) -- (24,4);\draw[very thin, color=black!50] (0,4-0.35) -- (0,4.35);\draw[very thin, color=black!50] (24,4-0.35) -- (24,4.35);   	
	\draw[very thin, color=black!50] (0,8) -- (24,8);\draw[very thin, color=black!50] (0,8-0.35) -- (0,8.35);\draw[very thin, color=black!50] (24,8-0.35) -- (24,8.35);
   	\node[black] (x0) at (1,8+2){\large{Machine C}};
		\draw[black, thin, ->] (3*2,8) to [out=45,in=135] (5*2,8);\node[black] (x5) at (4*2,8+1.2){\large{$x_{3,5,5,3}$}};
		\draw[black, thin, ->] (5*2,8) to [out=35,in=145] (8*2,8);\node[black] (x6) at (6.5*2,8+1.2){\large{$x_{5,8,6,3}$}};
		\draw[black, thin, ->] (8*2,8) to [out=45,in=135] (10*2,8);\node[black] (x4) at (9*2,8+1.2){\large{$x_{8,10,4,3}$}};
		\draw[gray, dashed, thin, ->] (2*2,8) to [out=-45,in=-135] (3*2,8);\node[black] (x0) at (2.5*2,8-1.2){\large{$x_{2,3,0,3}$}};
		\draw[gray, dashed, thin, ->] (10*2,8) to [out=-45,in=-135] (11*2,8);\node[black] (x0) at (10.5*2,8-1.2){\large{$x_{10,11,0,3}$}};
		\draw[gray, dashed, thin, ->] (11*2,8) to [out=-45,in=-135] (12*2,8);\node[black] (x0) at (11.5*2,8-1.2){\large{$x_{11,12,0,3}$}};
		
		\node (n2)  at (2*2 ,8-0.5){$2$}; \draw[color=gray] (2*2 ,8+0.2) -- (2*2 ,8-0.2);
		\node (n3)  at (3*2 ,8-0.5){$3$}; \draw[color=gray] (3*2 ,8+0.2) -- (3*2 ,8-0.2);
		\node (n5)  at (5*2 ,8-0.5){$5$}; \draw[color=gray] (5*2 ,8+0.2) -- (5*2 ,8-0.2);
		\node (n8)  at (8*2 ,8-0.5){$8$}; \draw[color=gray] (8*2 ,8+0.2) -- (8*2 ,8-0.2);
		\node (n10)  at (10*2 ,8-0.5){$10$}; \draw[color=gray] (10*2 ,8+0.2) -- (10*2 ,8-0.2);
		\node (n11)  at (11*2 ,8-0.5){$11$}; \draw[color=gray] (11*2 ,8+0.2) -- (11*2 ,8-0.2);
		\node (n12)  at (12*2 ,8-0.5){$12$}; \draw[color=gray] (12*2 ,8+0.2) -- (12*2 ,8-0.2); 		
   	\node[black] (x0) at (1,4+2){\large{Machine D}};   	
		\draw[black, thin, ->] (1*2,4) to [out=45,in=135] (2*2,4);\node[black] (x7) at (1.5*2,4+1.2){\large{$x_{1,2,7,4}$}};
		\draw[black, thin, ->] (3*2,4) to [out=25,in=155] (7*2,4);\node[black] (x9) at (5*2,4+1.2){\large{$x_{3,7,9,4}$}};
		\draw[black, thin, ->] (7*2,4) to [out=35,in=145] (10*2,4);\node[black] (x8) at (8.5*2,4+1.2){\large{$x_{7,10,8,4}$}};
		\draw[gray, dashed, thin, ->] (2*2,4) to [out=-45,in=-135] (3*2,4);\node[black] (x0) at (2.5*2,4-1.2){\large{$x_{2,3,0,4}$}};
		\draw[gray, dashed, thin, ->] (10*2,4) to [out=-45,in=-135] (11*2,4);\node[black] (x0) at (10.5*2,4-1.2){\large{$x_{10,11,0,4}$}};
		\draw[gray, dashed, thin, ->] (11*2,4) to [out=-45,in=-135] (12*2,4);\node[black] (x0) at (11.5*2,4-1.2){\large{$x_{11,12,0,4}$}};

\node (n1)  at (1*2 ,4-0.5){$1$}; \draw[color=gray] (1*2 ,4+0.2) -- (1*2 ,4-0.2);		
\node (n2)  at (2*2 ,4-0.5){$2$}; \draw[color=gray] (2*2 ,4+0.2) -- (2*2 ,4-0.2);
\node (n3)  at (3*2 ,4-0.5){$3$}; \draw[color=gray] (3*2 ,4+0.2) -- (3*2 ,4-0.2);
\node (n5)  at (7*2 ,4-0.5){$7$}; \draw[color=gray] (7*2 ,4+0.2) -- (7*2 ,4-0.2);
\node (n10)  at (10*2 ,4-0.5){$10$}; \draw[color=gray] (10*2 ,4+0.2) -- (10*2 ,4-0.2);
\node (n11)  at (11*2 ,4-0.5){$11$}; \draw[color=gray] (11*2 ,4+0.2) -- (11*2 ,4-0.2);
\node (n12)  at (12*2 ,4-0.5){$12$}; \draw[color=gray] (12*2 ,4+0.2) -- (12*2 ,4-0.2); 
		
   	\node[black] (x0) at (1,0+2){\large{Machine E}};   	
		\draw[black, thin, ->] (2*2,0) to [out=45,in=135] (4*2,0);\node[black] (x1) at (3*2,0+1.2){\large{$x_{2,4,1,5}$}};
		\draw[black, thin, ->] (4*2,0) to [out=35,in=145] (7*2,0);\node[black] (x10) at (5.5*2,0+1.2){\large{$x_{4,7,10,5}$}};
		\draw[black, thin, ->] (7*2,0) to [out=45,in=135] (8*2,0);\node[black] (x2) at (7.5*2,0+1.2){\large{$x_{7,8,2,5}$}};
		\draw[black, thin, ->] (8*2,0) to [out=45,in=135] (10*2,0);\node[black] (x3) at (9*2,0+1.2){\large{$x_{8,10,3,5}$}};
		\draw[gray, dashed, thin, ->] (10*2,0) to [out=-45,in=-135] (11*2,0);\node[black] (x0) at (10.5*2,0-1.2){\large{$x_{10,11,0,5}$}};
		\draw[gray, dashed, thin, ->] (11*2,0) to [out=-45,in=-135] (12*2,0);\node[black] (x0) at (11.5*2,0-1.2){\large{$x_{11,12,0,5}$}};
   	
		\node (n2)  at (2*2 ,0-0.5){$2$}; \draw[color=gray] (2*2 ,0+0.2) -- (2*2 ,0-0.2);
		\node (n4)  at (4*2 ,0-0.5){$4$}; \draw[color=gray] (4*2 ,0+0.2) -- (4*2 ,0-0.2);
		\node (n5)  at (7*2 ,0-0.5){$7$}; \draw[color=gray] (7*2 ,0+0.2) -- (7*2 ,0-0.2);
		\node (n8)  at (8*2 ,0-0.5){$8$}; \draw[color=gray] (8*2 ,0+0.2) -- (8*2 ,0-0.2);
		\node (n10)  at (10*2 ,0-0.5){$10$}; \draw[color=gray] (10*2 ,0+0.2) -- (10*2 ,0-0.2);
		\node (n11)  at (11*2 ,0-0.5){$11$}; \draw[color=gray] (11*2 ,0+0.2) -- (11*2 ,0-0.2);
		\node (n12)  at (12*2 ,0-0.5){$12$}; \draw[color=gray] (12*2 ,0+0.2) -- (12*2 ,0-0.2); 
   	\end{tikzpicture}
   	\caption{$AF$ solution for toy example of Figure \ref{fig:example}}
 \label{Fig:example_AF}
\end{figure}
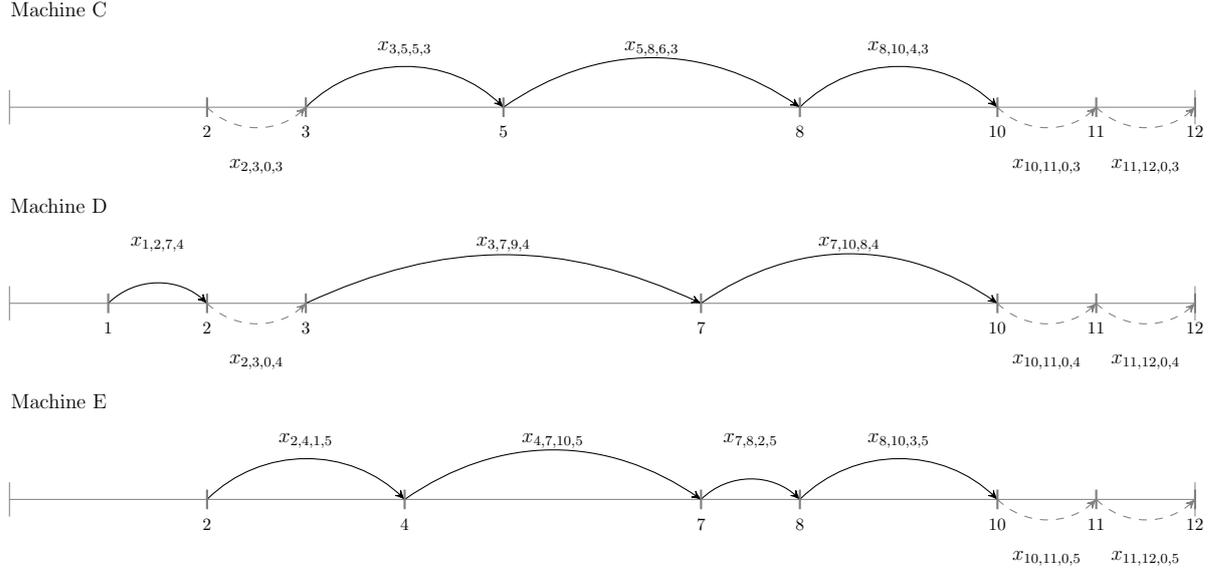

The objective function \eqref{FO:AF} seeks the minimization of the makespan. Constraints \eqref{constr1:AF} state that each job should be scheduled at least once. The flow conservation constraints are imposed by Constraints \eqref{constr2:AF}, while Constraints \eqref{constr3:AF} impose that the makespan must be greater or equal than the jobs' completion times. Constraints \eqref{constr4:AF} state that at most $p$ locations are selected, and Constraints \eqref{constr5:AF}--\eqref{constr8:AF} refer to the variables domain.
An AF representation for the solution depicted in Figure \ref{fig:example} is shown in Figure \ref{Fig:example_AF}.

\subsection{Column Generation}
\label{sec:CG}

The AF formulation \eqref{FO:AF}-\eqref{constr8:AF} models the DPMM ScheLoc problem by using a pseudo-polynomial number of variables and constraints, i.e., ${O}(nmT)$ and ${O}(mT)$, respectively. For instances where $n$, $m$ and $T$ are large, solving the AF model, or even its linear relaxation, may be computationally expensive. An alternative approach relies on the use of \emph{column generation} (CG) algorithms. CG algorithms are often employed to solve large scale \emph{linear programs} (LP) and embedded into branch-and-bound frameworks to solve mixed integer linear programs. CG methods make use of primal/dual information to iteratively solve LPs without enumerating all variables of the model. This is a noteworthy advantage of CG algorithms that, in practice, require less computer memory than solving the complete LP.

Briefly, CG algorithms consist of (i) solving a restricted LP containing only an initial subset of variables of the complete LP, (ii) obtaining the dual information associated to the optimal LP solution ($\lambda^*$), then (iii) using it to identify attractive columns with negative reduced costs (when minimization problem), and finally (iv) adding them to the restricted LP, (v) which is re-optimized. Steps (ii) to (v) are repeated until there is no variable with negative reduced cost, where the problem of finding attractive columns with negative reduced costs is known as the CG subproblem.

In order to solve the LP of AF formulation \eqref{FO:AF}-\eqref{constr8:AF} by means of CG, constraints \eqref{constr7:AF} and \eqref{constr8:AF} are relaxed and an initial LP, denoted by $LP(S)$, containing only a subset $S$ of variables from the original LP is considered. In our proposed CG algorithm, $LP(S)$ is initialized with variables $C_{\max}$, $y_{k}, k \in M$, $x_{qr0k}, (q,r,0,k) \in A$, and with all variables $x_{qrjk}$ that assume value one on an integer feasible solution $s$ (in Section \ref{sec:heuristics}, we discuss how to obtain feasible solutions for the DPMM ScheLoc problem). Then, at each iteration of the CG procedure, new attractive $x_{qrjk}$ variables (with negative reduced cost) are added to the $LP(S)$, where the reduced cost of variable $x_{qrjk}$ is given by:
\begin{align}
rc_{qrjk} =  - \pi_j - \tau_{kq} + \tau_{kr} - r\:\gamma_j\label{rc} &&& (q,r,j,k) \in A : j \in J, 
\end{align}
where the dual variables $\pi$, $\tau$ and $\gamma$ are associated with constraints \eqref{constr1:AF}, \eqref{constr2:AF} and \eqref{constr3:AF}, respectively. The proposed CG algorithm is summarized in Algorithm \ref{algo:CG}.
\vspace{-.1cm}
\begin{algorithm}
	\small
	{\bf Initialize $S$ considering a solution $s$;}{\footnotesize {\color{gray} \Comment $S$: set of initial columns}}\\
	{\bf Initialize $LP(S)$;}\\
	\Repeat{$S' = \emptyset$}{
		{\bf Solve} $LP(S)$ and obtain $\lambda^*$;{\footnotesize {\color{gray} \Comment $\lambda^*$: an optimal solution of $LP(S)$}}\\
		$(\pi,\tau,\gamma) \gets \: ${\bf Get\_duals($LP(S),\: \lambda^*$)};\\
		$S' \gets \emptyset$;\\
		\For{$(q,r,j,k) \in A \: | \: j \in J$ and $x_{qrjk} \notin S$}{
			$rc_{qrjk} \gets  - \pi_j - \tau_{kq} + \tau_{kr} - r\:\gamma_j$;{\footnotesize {\color{gray} \Comment Compute the reduced costs}}\\
			\If{$rc_{qrjk} < 0$}{
				$S' \gets S' \cup \{x_{qrjk}\}$
			}					
		}
		{\bf Select $S'' \subseteq S'$ according to a criterion;}{\footnotesize {\color{gray} \Comment $S''$: subset of columns to be added}}\\
		{\bf $S \gets S \cup S''$;}
	}
	{\bf return} $\lambda^*$;
	\caption{{\sc Column Generation}}
	\label{algo:CG}
\end{algorithm}

\section{Heuristic procedures}
\label{sec:heuristics}

As mentioned in the previous section, the size of the proposed AF formulation highly depends on the estimated upper bound $T$, i.e., the smaller the value of $T$, the smaller the number of variables and constraints. Moreover, a poor estimate of $T$ means unnecessary memory usage, limiting the capacity for solving large size instances. Therefore, obtaining good upper bounds becomes a primordial task when using AF formulations, especially for solving instances involving a large amount of jobs and candidate locations. To this aim, we developed three heuristic procedures that are able to generate high quality solutions in a reasonable computational time. 

\subsection{Integer programming-based heuristics}
\label{sec:ip_heur}

Two of the developed heuristics are based on \emph{mixed integer programming} (MIP), and consist in solving the AF formulation of Section \ref{sec:arcflow} considering a subset of variables and constraints. The first heuristic method, denoted by \texttt{AF\_CG}, initially executes the CG procedure described in Section \ref{sec:CG}, and then it solves the AF model \eqref{FO:AF}--\eqref{constr8:AF} containing only the subset of variables (columns) obtained by the CG algorithm. Since the CG procedure is initialized with the columns associated with a feasible solution, \texttt{AF\_CG} always returns a feasible integer solution to the DPMM ScheLoc problem.

The second MIP-based heuristic, referred to as \texttt{AF\_subsetM}, consists in solving the AF formulation \eqref{FO:AF}--\eqref{constr8:AF} considering only the variables and constraints related to a reduced subset $\overline{M}$ of candidate locations, with $p \leq |\overline{M}| \leq |M|$. On the one hand, when $\overline{M}$ is small (e.g., $|\overline{M}| = p$) the formulation will be smaller and easier to solve than the complete model. However, the choice of the elements that will compose the set $\overline{M}$ should be accurate, otherwise the quality of the solution may be compromised. On the other hand, by increasing the set $\overline{M}$, the chance of choosing a set of candidate locations that makes it possible to find an optimal solution to the complete model increases, but solving the resulting formulation becomes more time consuming. Thus, \texttt{AF\_subsetM} requires a procedure to select the $p + \delta$ candidate locations, with $0 \leq \delta \leq |M| - p$, where $\delta$ is a parameter that controls the solution space and the time required to solve the reduced formulation.

\subsection{Iterated Local Search}
\label{sec:ils}

The third heuristic method is an algorithm based on the \emph{iterated local search} (ILS) metaheuristic, and its description is presented next.

ILS is a simple and effective metaheuristic that has been successfully applied to solve several optimization problems, including scheduling and location problems (see, e.g., \citealt{Lourenco2019}). The ILS consists of four main components: (i) a constructive method to generate an initial solution; (ii) a local search procedure to explore solutions in a reduced search space (neighborhood); (iii) a perturbation mechanism to escape from local optimal solutions; and (iv) an acceptance criterion to decide whether the incumbent solution should be replaced by a neighbor (for more details, we refer the reader to \citealt{Lourenco2019}). In our implementation, the ILS is embedded into a multi-start scheme that invokes it $n_\text{iter}$ times, as shown in Algorithm \ref{algo:MS-ILS}. 

\begin{algorithm}
	\caption{{\sc Multi-Start Iterated Local Search}}
	\label{algo:MS-ILS}
	$s_\text{best} \leftarrow \emptyset$; $f(s_\text{best}) \leftarrow \infty$; \\
	\For{$iter = 1$ to $n_\text{iter}$}
	{
		$s_i \leftarrow s \leftarrow \texttt{GenerateInitialSolution()}$; \\
		$iterILS \leftarrow 1$; \\
		\While{$iterILS \leq n_\text{ils}$}
		{
			$s \leftarrow \texttt{LocalSearch($s$)}$; \\ 
			\If({{\color{gray}\Comment {\footnotesize Acceptance Criterion}}}){$f(s) < f(s_i)$}
			{
				$s_i \leftarrow s$; $iterILS \leftarrow 0$; \\
			}
			$s \leftarrow \texttt{Perturbation($s_i$)}$; \\
			$iterILS \leftarrow iterILS + 1$; \\
		}
		\If{$f(s_i) < f(s_\text{best})$}
		{
			$s_\text{best} \leftarrow s$; \\
		}
	}
	{\bf return} $s_\text{best}$;
\end{algorithm}

To generate an initial solution, we adopt a sequential approach that first selects $p$ machine locations, then it assigns the jobs to the machines (locations) and solves the scheduling. As it is a multi-start ILS, the initial solution must consider a random component in order to avoid obtaining the same initial solution on all $n_\text{iter}$ iterations. In this way, for the first $n_\text{iter}-1$ iterations the locations are chosen randomly, while in the last iteration they are selected by a deterministic algorithm that works as follows. 
Let $\kappa_j = \argmin_{k \in M} \{r_{jk}\}$ (or, if $N=M$, $\kappa_j = \argmin_{k \in M:k \neq j} \{r_{jk}\}$) be the closest machine location of job $j$, and $r_j = r_{j,\kappa_j}$ be the release date of job $j$ for location $\kappa_j$. Initially, for each $j \in J$, the algorithm stores the pair $(\kappa_j, r_j)$ in a vector $vec$, and sort them in non-decreasing order of release dates. Then, it chooses the first $p$ distinct locations of sorted $vec$. If the number of chosen locations is less than $p$, the procedure is repeated to select the remaining ones, but now disregarding the already chosen locations. Then, the algorithm assigns jobs to the machines in the selected locations. 

In order to explain the assignment strategy, let $M'$ be the set of selected locations, $LJ$ be a list of unassigned jobs, and $\bar{s}$ be an incomplete solution (under construction). For each chosen location $k \in M'$, the job $j' \in LJ$ whose processing can be started as early as possible is identified (i.e., considering the job release dates and the machine completion times in $\bar{s}$), and the pair $(k,j')$ is inserted in a list $CL$ of candidate assignments (ties are broken by choosing the job with largest processing time). Then, the assignment $(k^*,j^*) \in CL$ leading to the earliest completion time is performed, $j^*$ is removed from $LJ$, and $CL$ is cleared. The procedure restarts until $LJ$ becomes empty. Once all the assignments are done, the scheduling of jobs on each machine is solved by using the ERD rule. 

\begin{figure}[htbp]
	\centering
	\includegraphics[scale=0.35]{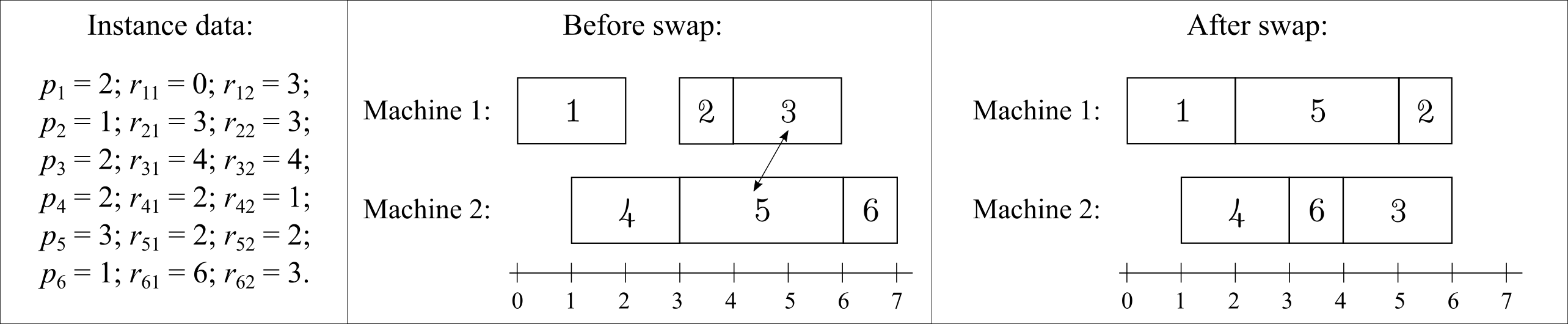}
	\caption{Example of a swap move.}
	\label{fig:swap}
\end{figure}

After obtaining the initial solution, a local search is performed by applying swap moves involving two jobs assigned to two different machines. Instead of simply exchanging the job positions, we also move them to the best position in the new machines, such that the resulting schedule still respects the ERD rule, as illustrated in Figure \ref{fig:swap}. To make local search efficient, we use an auxiliary data structure that stores, for each job $j \in J$, the best position it can be scheduled in each machine $k \in M'$. In addition, for each subsequence $\sigma$ of consecutive jobs in the current solution, we store its total processing time $P(\sigma)$, and the earliest time it can start to be processed (in the machine they are scheduled), $E(\sigma)$. For a subsequence $\sigma_j^k$ involving a single job $j$ scheduled on machine $k$, $P(\sigma_j^k)$ and $E(\sigma_j^k)$ are equal to $p_j$ and $r_{jk}$, respectively. For larger subsequences, they are computed by concatenation $\oplus$ of smaller subsequences, according to equations \eqref{eq:P} and \eqref{eq:E} (see, e.g., \citealt{Vidal2013}).

\begin{align}
P(\sigma' \oplus \sigma'') = P(\sigma') + P(\sigma'') \label{eq:P} \\ 
E(\sigma' \oplus \sigma'') = \max\{E(\sigma'') - P(\sigma'), E(\sigma')\} \label{eq:E}
\end{align}

\vspace{-0.25cm}

By using the auxiliary data structures mentioned above, the costs of the solutions evaluated during the local search can be computed in constant time. In order to illustrate how to use this data structure, let, for a given solution, $\rho_{jk}$ represent the current position of job $j$ on the schedule of machine $k$, $\rho'_{jm}$ be the best position in which the job $j$ could be inserted in the schedule of machine $m$, and $\sigma_j^k$ be a subsequence containing only the job $j$ in machine $k$.

Let us now consider the swap move between jobs $j$ and $l$, initially assigned to machines $k$ and $m$, respectively. In order to obtain the new completion times of machines $k$ and $m$, a series of concatenations should be performed. The concatenations associated with machine $m$ are executed first.

\begin{figure*}[!h]
	\begin{framed}
		\centering
		\begin{subfigure}[b]{0.243\textwidth}
			\centering
			\begin{tikzpicture}[scale=0.90]
  \node [text width=0.05cm,align=center] at (-0.35,0.25) {\scriptsize{$k$:}};
  \node [text width=0.05cm,align=center] at (-0.45,-0.75) {\scriptsize{$m$:}};
  \draw[draw=black] (0,0) rectangle ++(1.5,.5) node[pos=.5] {$\sigma_1$};
  \draw[draw=black,fill=blue!14] (1.5,0) rectangle ++(.5,.5) node[pos=.5] {$j$};
  \draw[draw=black] (2.0,0) rectangle ++(1.5,.5) node[pos=.5] {$\sigma_2$};
  \draw[draw=black,fill=blue!14] (0,-1) rectangle ++(0.5,.5) node[pos=.5] {$l$};
  \draw[draw=black] (0.5,-1) rectangle ++(3,.5) node[pos=.5] {$\sigma_3$};
  \draw [->,-latex] (1.75,-0.05) -- (0.0,-0.45);
  \node [text width=3.5cm,align=center] at (1.75, -1.5) {\scriptsize{$\sigma(m)=\sigma_j^m \oplus \sigma_3$}};
\end{tikzpicture}
			\vspace{-0.65cm}
			\caption{\scriptsize Case 1}
		\end{subfigure}
		\begin{subfigure}[b]{0.243\textwidth}
			\centering
			\begin{tikzpicture}[scale=0.90]
  \node [text width=0.05cm,align=center] at (-0.35,0.25) {\scriptsize{$k$:}};
  \node [text width=0.05cm,align=center] at (-0.45,-0.75) {\scriptsize{$m$:}};
  \draw[draw=black] (0,0) rectangle ++(1.5,.5) node[pos=.5] {$\sigma_1$};
  \draw[draw=black,fill=blue!14] (1.5,0) rectangle ++(.5,.5) node[pos=.5] {$j$};
  \draw[draw=black] (2.0,0) rectangle ++(1.5,.5) node[pos=.5] {$\sigma_2$};
  \draw[draw=black] (0,-1) rectangle ++(3,.5) node[pos=.5] {$\sigma_3$};
  \draw[draw=black,fill=blue!14] (3,-1) rectangle ++(0.5,.5) node[pos=.5] {$l$};
  \draw [->,-latex] (1.75,-0.05) -- (0,-0.45);
  \node [text width=3.5cm,align=center] at (1.75, -1.5) {\scriptsize{$\sigma(m)= \sigma_j^m \oplus \sigma_3$}};
\end{tikzpicture}
			\vspace{-0.65cm}
			\caption{\scriptsize Case 2}
		\end{subfigure}
		\begin{subfigure}[b]{0.243\textwidth}
			\centering
			\begin{tikzpicture}[scale=0.90]
  \node [text width=0.05cm,align=center] at (-0.35,0.25) {\scriptsize{$k$:}};
  \node [text width=0.05cm,align=center] at (-0.45,-0.75) {\scriptsize{$m$:}};
  \draw[draw=black] (0,0) rectangle ++(1.5,.5) node[pos=.5] {$\sigma_1$};
  \draw[draw=black,fill=blue!14] (1.5,0) rectangle ++(.5,.5) node[pos=.5] {$j$};
  \draw[draw=black] (2.0,0) rectangle ++(1.5,.5) node[pos=.5] {$\sigma_2$};
  \draw[draw=black] (0,-1) rectangle ++(3,.5) node[pos=.5] {$\sigma_3$};
  \draw[draw=black,fill=blue!14] (3,-1) rectangle ++(0.5,.5) node[pos=.5] {$l$};
  \draw [->,-latex] (1.75,-0.05) -- (3,-0.45);
  \node [text width=3.5cm,align=center] at (1.75, -1.5) {\scriptsize{$\sigma(m)= \sigma_3 \oplus \sigma_j^m$}};
\end{tikzpicture}
			\vspace{-0.65cm}
			\caption{\scriptsize Case 3}
		\end{subfigure}
		\begin{subfigure}[b]{0.243\textwidth}
			\centering
			\begin{tikzpicture}[scale=0.90]
  \node [text width=0.05cm,align=center] at (-0.35,0.25) {\scriptsize{$k$:}};
  \node [text width=0.05cm,align=center] at (-0.45,-0.75) {\scriptsize{$m$:}};
  \draw[draw=black] (0,0) rectangle ++(1.5,.5) node[pos=.5] {$\sigma_1$};
  \draw[draw=black,fill=blue!14] (1.5,0) rectangle ++(.5,.5) node[pos=.5] {$j$};
  \draw[draw=black] (2.0,0) rectangle ++(1.5,.5) node[pos=.5] {$\sigma_2$};
  \draw[draw=black] (0,-1) rectangle ++(1.5,.5) node[pos=.5] {$\sigma_3$};
  \draw[draw=black] (1.5,-1) rectangle ++(1.5,.5) node[pos=.5] {$\sigma_4$};
  \draw[draw=black,fill=blue!14] (3,-1) rectangle ++(0.5,.5) node[pos=.5] {$l$};
  \draw [->,-latex] (1.75,-0.05) -- (1.5,-0.45);
  \node [text width=3.5cm,align=center] at (1.75, -1.5) {\scriptsize{$\sigma(m)=\sigma_3 \oplus \sigma_j^m \oplus \sigma_4$}};
\end{tikzpicture}
			\vspace{-0.65cm}
			\caption{\scriptsize Case 4}
		\end{subfigure}\\ 
		\vspace{0.5cm}
		\begin{subfigure}[b]{0.245\textwidth}
			\centering
			\begin{tikzpicture}[scale=0.90]
  \node [text width=0.05cm,align=center] at (-0.35,0.25) {\scriptsize{$k$:}};
  \node [text width=0.05cm,align=center] at (-0.45,-0.75) {\scriptsize{$m$:}};
  \draw[draw=black] (0,0) rectangle ++(1.5,.5) node[pos=.5] {$\sigma_1$};
  \draw[draw=black,fill=blue!14] (1.5,0) rectangle ++(.5,.5) node[pos=.5] {$j$};
  \draw[draw=black] (2.0,0) rectangle ++(1.5,.5) node[pos=.5] {$\sigma_2$};
  \draw[draw=black] (0,-1) rectangle ++(1.5,.5) node[pos=.5] {$\sigma_3$};
  \draw[draw=black,fill=blue!14] (1.5,-1) rectangle ++(.5,.5) node[pos=.5] {$l$};
  \draw[draw=black] (1.5,-1) rectangle ++(2,.5) node[pos=.5] {$\sigma_4$};
  \draw [->,-latex] (1.75,-0.05) -- (0,-0.45);
  \node [text width=3.5cm,align=center] at (1.75, -1.5) {\scriptsize{$\sigma(m)=\sigma_j^m \oplus \sigma_3 \oplus \sigma_4$}};
\end{tikzpicture}
			\vspace{-0.65cm}
			\caption{\scriptsize Case 5}
		\end{subfigure}
		\begin{subfigure}[b]{0.245\textwidth}
			\centering
			\begin{tikzpicture}[scale=0.90]
  \node [text width=0.05cm,align=center] at (-0.35,0.25) {\scriptsize{$k$:}};
  \node [text width=0.05cm,align=center] at (-0.45,-0.75) {\scriptsize{$m$:}};
  \draw[draw=black] (0,0) rectangle ++(1.5,.5) node[pos=.5] {$\sigma_1$};
  \draw[draw=black,fill=blue!14] (1.5,0) rectangle ++(.5,.5) node[pos=.5] {$j$};
  \draw[draw=black] (2.0,0) rectangle ++(1.5,.5) node[pos=.5] {$\sigma_2$};
  \draw[draw=black] (0,-1) rectangle ++(1,.5) node[pos=.5] {$\sigma_4$};
  \draw[draw=black] (1,-1) rectangle ++(1,.5) node[pos=.5] {$\sigma_5$};
  \draw[draw=black,fill=blue!14] (2,-1) rectangle ++(.5,.5) node[pos=.5] {$l$};
  \draw[draw=black] (2.5,-1) rectangle ++(1,.5) node[pos=.5] {$\sigma_3$};
  \draw [->,-latex] (1.75,-0.05) -- (1,-0.45);
  \node [text width=3.85cm,align=center] at (1.75, -1.5) {\scriptsize{$\sigma(m)=\sigma_4 \oplus \sigma_j^m \oplus \sigma_5 \oplus \sigma_3$}};
\end{tikzpicture}
			\vspace{-0.65cm}
			\caption{\scriptsize Case 6}
		\end{subfigure}
		\begin{subfigure}[b]{0.245\textwidth}
			\centering
			\begin{tikzpicture}[scale=0.90]
  \node [text width=0.05cm,align=center] at (-0.35,0.25) {\scriptsize{$k$:}};
  \node [text width=0.05cm,align=center] at (-0.45,-0.75) {\scriptsize{$m$:}};
  \draw[draw=black] (0,0) rectangle ++(1.5,.5) node[pos=.5] {$\sigma_1$};
  \draw[draw=black,fill=blue!14] (1.5,0) rectangle ++(.5,.5) node[pos=.5] {$j$};
  \draw[draw=black] (2.0,0) rectangle ++(1.5,.5) node[pos=.5] {$\sigma_2$};
  \draw[draw=black] (0,-1) rectangle ++(2,.5) node[pos=.5] {$\sigma_3$};
  \draw[draw=black,fill=blue!14] (2,-1) rectangle ++(.5,.5) node[pos=.5] {$l$};
  \draw[draw=black] (2.5,-1) rectangle ++(1,.5) node[pos=.5] {$\sigma_4$};
  \draw [->,-latex] (1.75,-0.05) -- (2,-0.45);
  \node [text width=3.5cm,align=center] at (1.75, -1.5) {\scriptsize{$\sigma(m)=\sigma_3 \oplus \sigma_j^m \oplus \sigma_4$}};
\end{tikzpicture}
			\vspace{-0.65cm}
			\caption{\scriptsize Case 7}
		\end{subfigure}
	\end{framed}
	\vspace{-0.5cm}
	\caption{Concatenation cases when moving $j$ to a position before $l$.}
	\label{fig:concatbefore}
\end{figure*}
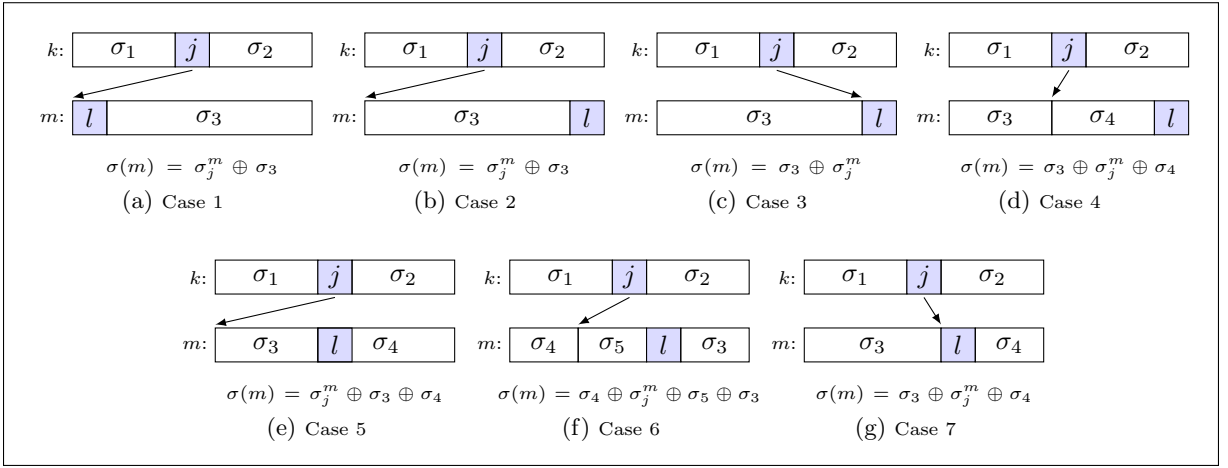

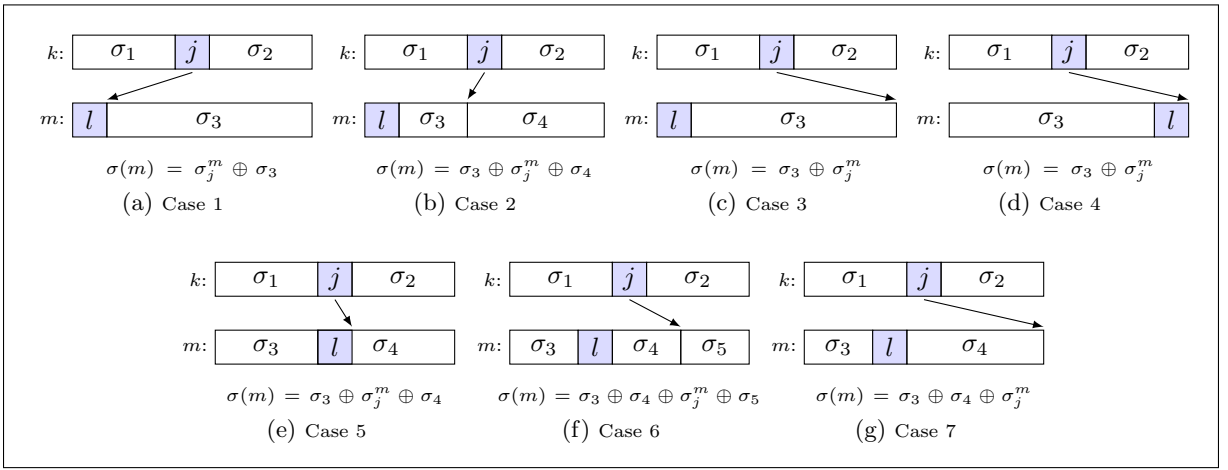
\begin{figure*}[!h]
	\begin{framed}
		\centering
		\begin{subfigure}[b]{0.243\textwidth}
			\centering
			\begin{tikzpicture}[scale=0.90]
  \node [text width=0.05cm,align=center] at (-0.35,0.25) {\scriptsize{$k$:}};
  \node [text width=0.05cm,align=center] at (-0.45,-0.75) {\scriptsize{$m$:}};
  \draw[draw=black] (0,0) rectangle ++(1.5,.5) node[pos=.5] {$\sigma_1$};
  \draw[draw=black,fill=blue!14] (1.5,0) rectangle ++(.5,.5) node[pos=.5] {$j$};
  \draw[draw=black] (2.0,0) rectangle ++(1.5,.5) node[pos=.5] {$\sigma_2$};
  \draw[draw=black,fill=blue!14] (0,-1) rectangle ++(0.5,.5) node[pos=.5] {$l$};
  \draw[draw=black] (0.5,-1) rectangle ++(3,.5) node[pos=.5] {$\sigma_3$};
  \draw [->,-latex] (1.75,-0.05) -- (0.5,-0.45);
  \node [text width=3.5cm,align=center] at (1.75, -1.5) {\scriptsize{$\sigma(m)=\sigma_j^m \oplus \sigma_3$}};
\end{tikzpicture}
			\vspace{-0.65cm}
			\caption{\scriptsize Case 1}
		\end{subfigure}
		\begin{subfigure}[b]{0.243\textwidth}
			\centering
			\begin{tikzpicture}[scale=0.90]
  \node [text width=0.05cm,align=center] at (-0.35,0.25) {\scriptsize{$k$:}};
  \node [text width=0.05cm,align=center] at (-0.45,-0.75) {\scriptsize{$m$:}};
  \draw[draw=black] (0,0) rectangle ++(1.5,.5) node[pos=.5] {$\sigma_1$};
  \draw[draw=black,fill=blue!14] (1.5,0) rectangle ++(.5,.5) node[pos=.5] {$j$};
  \draw[draw=black] (2.0,0) rectangle ++(1.5,.5) node[pos=.5] {$\sigma_2$};
  \draw[draw=black,fill=blue!14] (0,-1) rectangle ++(0.5,.5) node[pos=.5] {$l$};
  \draw[draw=black] (0.5,-1) rectangle ++(1,.5) node[pos=.5] {$\sigma_3$};
  \draw[draw=black] (1.5,-1) rectangle ++(2,.5) node[pos=.5] {$\sigma_4$};
  \draw [->,-latex] (1.75,-0.05) -- (1.5,-0.45);
  \node [text width=3.5cm,align=center] at (1.75, -1.5) {\scriptsize{$\sigma(m)=\sigma_3 \oplus \sigma_j^m \oplus \sigma_4$}};
\end{tikzpicture}
			\vspace{-0.65cm}
			\caption{\scriptsize Case 2}
		\end{subfigure}
		\begin{subfigure}[b]{0.243\textwidth}
			\centering
			\begin{tikzpicture}[scale=0.90]
  \node [text width=0.05cm,align=center] at (-0.35,0.25) {\scriptsize{$k$:}};
  \node [text width=0.05cm,align=center] at (-0.45,-0.75) {\scriptsize{$m$:}};
  \draw[draw=black] (0,0) rectangle ++(1.5,.5) node[pos=.5] {$\sigma_1$};
  \draw[draw=black,fill=blue!14] (1.5,0) rectangle ++(.5,.5) node[pos=.5] {$j$};
  \draw[draw=black] (2.0,0) rectangle ++(1.5,.5) node[pos=.5] {$\sigma_2$};
  \draw[draw=black,fill=blue!14] (0,-1) rectangle ++(0.5,.5) node[pos=.5] {$l$};
  \draw[draw=black] (0.5,-1) rectangle ++(3,.5) node[pos=.5] {$\sigma_3$};
  \draw [->,-latex] (1.75,-0.05) -- (3.5,-0.45);
  \node [text width=3.5cm,align=center] at (1.75, -1.5) {\scriptsize{$\sigma(m)=\sigma_3 \oplus \sigma_j^m$}};
\end{tikzpicture}
			\vspace{-0.65cm}
			\caption{\scriptsize Case 3}
		\end{subfigure}
		\begin{subfigure}[b]{0.243\textwidth}
			\centering
			\begin{tikzpicture}[scale=0.90]
  \node [text width=0.05cm,align=center] at (-0.35,0.25) {\scriptsize{$k$:}};
  \node [text width=0.05cm,align=center] at (-0.45,-0.75) {\scriptsize{$m$:}};
  \draw[draw=black] (0,0) rectangle ++(1.5,.5) node[pos=.5] {$\sigma_1$};
  \draw[draw=black,fill=blue!14] (1.5,0) rectangle ++(.5,.5) node[pos=.5] {$j$};
  \draw[draw=black] (2.0,0) rectangle ++(1.5,.5) node[pos=.5] {$\sigma_2$};
  \draw[draw=black] (0,-1) rectangle ++(3,.5) node[pos=.5] {$\sigma_3$};
  \draw[draw=black,fill=blue!14] (3,-1) rectangle ++(0.5,.5) node[pos=.5] {$l$};
  \draw [->,-latex] (1.75,-0.05) -- (3.5,-0.45);
  \node [text width=3.5cm,align=center] at (1.75, -1.5) {\scriptsize{$\sigma(m)= \sigma_3 \oplus \sigma_j^m$}};
\end{tikzpicture}
			\vspace{-0.65cm}
			\caption{\scriptsize Case 4}
		\end{subfigure}\\ 
		\vspace{0.5cm}
		\begin{subfigure}[b]{0.245\textwidth}
			\centering
			\begin{tikzpicture}[scale=0.90]
  \node [text width=0.05cm,align=center] at (-0.35,0.25) {\scriptsize{$k$:}};
  \node [text width=0.05cm,align=center] at (-0.45,-0.75) {\scriptsize{$m$:}};
  \draw[draw=black] (0,0) rectangle ++(1.5,.5) node[pos=.5] {$\sigma_1$};
  \draw[draw=black,fill=blue!14] (1.5,0) rectangle ++(.5,.5) node[pos=.5] {$j$};
  \draw[draw=black] (2.0,0) rectangle ++(1.5,.5) node[pos=.5] {$\sigma_2$};
  \draw[draw=black] (0,-1) rectangle ++(1.5,.5) node[pos=.5] {$\sigma_3$};
  \draw[draw=black,fill=blue!14] (1.5,-1) rectangle ++(.5,.5) node[pos=.5] {$l$};
  \draw[draw=black] (1.5,-1) rectangle ++(2,.5) node[pos=.5] {$\sigma_4$};
  \draw [->,-latex] (1.75,-0.05) -- (2,-0.45);
  \node [text width=3.5cm,align=center] at (1.75, -1.5) {\scriptsize{$\sigma(m)=\sigma_3 \oplus \sigma_j^m \oplus \sigma_4$}};
\end{tikzpicture}
			\vspace{-0.65cm}
			\caption{\scriptsize Case 5}
		\end{subfigure}
		\begin{subfigure}[b]{0.245\textwidth}
			\centering
			\begin{tikzpicture}[scale=0.90]
  \node [text width=0.05cm,align=center] at (-0.35,0.25) {\scriptsize{$k$:}};
  \node [text width=0.05cm,align=center] at (-0.45,-0.75) {\scriptsize{$m$:}};
  \draw[draw=black] (0,0) rectangle ++(1.5,.5) node[pos=.5] {$\sigma_1$};
  \draw[draw=black,fill=blue!14] (1.5,0) rectangle ++(.5,.5) node[pos=.5] {$j$};
  \draw[draw=black] (2.0,0) rectangle ++(1.5,.5) node[pos=.5] {$\sigma_2$};
  \draw[draw=black] (0,-1) rectangle ++(1,.5) node[pos=.5] {$\sigma_3$};
  \draw[draw=black,fill=blue!14] (1,-1) rectangle ++(.5,.5) node[pos=.5] {$l$};
  \draw[draw=black] (1.5,-1) rectangle ++(1,.5) node[pos=.5] {$\sigma_4$};
  \draw[draw=black] (2.5,-1) rectangle ++(1,.5) node[pos=.5] {$\sigma_5$};
  \draw [->,-latex] (1.75,-0.05) -- (2.5,-0.45);
  \node [text width=3.85cm,align=center] at (1.75, -1.5) {\scriptsize{$\sigma(m)=\sigma_3 \oplus \sigma_4 \oplus \sigma_j^m \oplus \sigma_5$}};
\end{tikzpicture}
			\vspace{-0.65cm}
			\caption{\scriptsize Case 6}
		\end{subfigure}
		\begin{subfigure}[b]{0.245\textwidth}
			\centering
			\begin{tikzpicture}[scale=0.90]
  \node [text width=0.05cm,align=center] at (-0.35,0.25) {\scriptsize{$k$:}};
  \node [text width=0.05cm,align=center] at (-0.45,-0.75) {\scriptsize{$m$:}};
  \draw[draw=black] (0,0) rectangle ++(1.5,.5) node[pos=.5] {$\sigma_1$};
  \draw[draw=black,fill=blue!14] (1.5,0) rectangle ++(.5,.5) node[pos=.5] {$j$};
  \draw[draw=black] (2.0,0) rectangle ++(1.5,.5) node[pos=.5] {$\sigma_2$};
  \draw[draw=black] (0,-1) rectangle ++(1,.5) node[pos=.5] {$\sigma_3$};
  \draw[draw=black,fill=blue!14] (1,-1) rectangle ++(.5,.5) node[pos=.5] {$l$};
  \draw[draw=black] (1.5,-1) rectangle ++(2,.5) node[pos=.5] {$\sigma_4$};
  \draw [->,-latex] (1.75,-0.05) -- (3.5,-0.45);
  \node [text width=3.5cm,align=center] at (1.75, -1.5) {\scriptsize{$\sigma(m)=\sigma_3 \oplus \sigma_4 \oplus \sigma_j^m$}};
\end{tikzpicture}
			\vspace{-0.65cm}
			\caption{\scriptsize Case 7}
		\end{subfigure}
	\end{framed}
	\vspace{-0.5cm}
	\caption{Concatenation cases when moving $j$ to a position after $l$.}
	\label{fig:concatafter}
\end{figure*}

Depending on the values of $\rho_{lm}$ and $\rho'_{jm}$ the concatenations should be performed according to one of the fourteen possible cases (illustrated in Figures \ref{fig:concatbefore} and \ref{fig:concatafter}). If job $j$ is moved to a position before the current position of job $l$ (i.e., $\rho'_{jm} < \rho_{lm}$), the concatenations are performed as shown in Figure \ref{fig:concatbefore}. Otherwise, if $j$ is moved to a position after the current position of job $l$ (i.e., $\rho'_{jm} > \rho_{lm}$), the concatenations should be performed as shown in Figure \ref{fig:concatafter}. Then, the concatenations associated to the machine $k$ are computed similarly. The local search evaluates all solutions that could be obtained by swapping every pair of jobs scheduled in distinct machines and executes the best improving move at the end. This procedure is repeated until no improving move exists.

Finally, after finding a local optimal solution, the perturbation procedure is executed. It randomly exchanges each selected location $k \in M'$ with a non-selected location $m \in M \setminus M'$, iteratively. Basically, from $i=1, \dots, p$, the schedule associated to the $i^ \text{th}$ chosen location is moved to a new random location from $M\setminus M'$, and the set $M'$ is updated. After exchanging the selected locations, the schedule of each machine is updated according to the ERD rule.

\section{Exact framework}
\label{sec:exactframework}

By using the CG method of Section \ref{sec:CG} and the heuristic procedures of Section \ref{sec:heuristics} it is possible to compute the optimality gap and 
occasionally prove the optimality of the solutions without the need to solve the AF model \eqref{FO:AF}-\eqref{constr8:AF}. 
In view of this, we embedded these methods in a framework algorithm that executes each one individually and updates the bounds after each execution. The order in which the procedures are executed was defined according to their expected running times, i.e., starting from the least time-consuming. In order to avoid a slow convergence of the algorithm, all formulations are solved for a given time limit. Since the execution of the column generation method may be computationally expensive, an initial lower bound is computed as shown in Equation \eqref{eq:initLB}. 
\begin{align}
LB = \left\lceil\left(\sum\nolimits_{j \in N} p_i / p\right) \right\rceil + \min_{j \in N, k \in M} \{r_{jk}\} \label{eq:initLB}
\end{align}

The resulting framework algorithm is presented in Algorithm \ref{algo:exact_scheloc}. First, initial lower and upper bounds are obtained by Equation \eqref{eq:initLB} and by the ILS presented in Section \ref{sec:ils} (Lines \ref{algo1:initLB} and \ref{algo1:ILS}), respectively. If they are equal, the ILS solution is optimal, and the algorithm terminates. Otherwise, seeking to improve the \emph{lower bound} ($LB$), it runs the CG procedure of Section \ref{sec:CG}, starting with the columns associated with the ILS solution (Line \ref{algo1:CG}). In case the $LB$ is improved, and it is equal to the \emph{upper bound} ($UB$), the optimality of the ILS solution is proved. If not, the framework tries to find a better integer solution by executing the \texttt{AF\_CG} heuristic, i.e., by solving the AF formulation considering only the variables generated by the CG procedure. Then, a local search (presented in Section \ref{sec:ils}) is performed on the obtained solution, and the $UB$ is updated (Lines \ref{algo1:MIPCG}-\ref{algo1:MIPCG-UB2}). If the optimality gap is not closed (i.e., $UB > LB$), the \texttt{AF\_subsetM} heuristic is invoked, i.e., the AF formulation considering only a subset of candidate locations is solved (e.g., with $\overline{M}=M''\cup R$, where $M''$ contains the $p$ machine locations chosen in the current best solution, and $R$ contains $\min\{\lfloor0.5p\rfloor, m-p\}$ locations randomly chosen from $M \setminus M''$), and, again, a local search is executed and the $UB$ is updated (Lines \ref{algo1:AFSubset}-\ref{algo1:AFSubset-UB2}). Finally, if the optimal solution is still not found (or proved), the full AF model is solved using the current best solution as a warm start, followed by another local search, if the time limit is reached (Lines \ref{algo1:AFFull}-\ref{algo1:AFFull-UB}).

\begin{algorithm}[h]
	\caption{{\sc Exact Framework for the ScheLoc Problem}}
	\label{algo:exact_scheloc}
	$opt \leftarrow false$; \\
	$LB \leftarrow \texttt{computeLB()}$; \label{algo1:initLB}\\
	$s \leftarrow \texttt{ILS()}$; $UB \leftarrow f(s)$; \label{algo1:ILS}\\
	\If{$(LB < UB)$}
	{
		$LB', cols \leftarrow \texttt{ColumnGeneration($s$)}$; \label{algo1:CG}\\
		$LB \leftarrow \max\{LB, LB'\}$; \\
		$\textbf{if }(LB = UB) \textbf{ then break}$; \\
		$s' \leftarrow \texttt{LocalSearch($\texttt{solveAF\_CG($cols, timelimit$)}$)}$; \label{algo1:MIPCG}\\
		\If{$(f(s') < f(s))$} %
		{\label{algo1:MIPCG-UB1}
			$UB \leftarrow f(s')$; $s \leftarrow s'$; \\
			$\textbf{if }(LB = UB) \textbf{ then break}$; \label{algo1:MIPCG-UB2}\\
		}
		$s' \leftarrow \texttt{LocalSearch($\texttt{solveAF\_SubsetM($M'', timelimit$)}$)}$; \label{algo1:AFSubset}\\
		\If{$(f(s') < f(s))$}
		{\label{algo1:AFSubset-UB1}
			$UB \leftarrow f(s')$; $s \leftarrow s'$; \\
			$\textbf{if }(LB = UB) \textbf{ then break}$; \label{algo1:AFSubset-UB2}\\
		}
		$s' \leftarrow \texttt{LocalSearch($\texttt{solveAF\_Full($s, timelimit$)}$)}$; \label{algo1:AFFull}\\
		\If{$(f(s') < f(s))$}
		{
			$UB \leftarrow f(s')$; $s \leftarrow s'$; \label{algo1:AFFull-UB}\\
		}
	}
	$\textbf{if }(LB = UB) \textbf{ then } opt \leftarrow true$; \\
	{\bf return} $s, opt$; \\
\end{algorithm}

\section{Computational results}
\label{sec:results}

We conducted extensive computational experiments to evaluate the performance of the proposed methods. The algorithms were coded in C++ and executed on a single thread of a computer equipped with an Intel Core i5-5200U processor with 2.20GHz and 16 GB of RAM, running under Linux Mint 17.2 64-bit operating system. Gurobi Optimizer 8.1 was adopted to solve the mathematical formulations. 

\subsection{Benchmark instances}
\label{sec:instances}

Four sets of instances proposed by \citet{Hessler2017} were considered. The first two sets are composed of randomly generated instances, the third set is composed of instances derived from a randomly generated network, and the fourth set contains instances with machine candidate locations randomly chosen from a square plane. All instances are publicly available at \url{https://drive.google.com/open?id=0B_-MYN0r5mjqOEdGSEhoLW9Ubk0}. The main characteristics of each set of instances are presented in Table \ref{tab:instances}.

\begin{table}[h]
	\centering
	\footnotesize
	\caption{Instances characteristics}
	\begin{tabular}{ccccccccc}
		\toprule
		set & name & structure & \#inst. & $n$ & $m$ & $p$ & $r_{jk}$ values & $p_{j}$ values \\ \midrule
		1 & small & random, with $|J|\neq|M|$ & 50 & $[2, 30]$ & $[2, 10]$ & $[1, 9]$ & integer & integer \\ 
		2 & large & random, with $|J|\neq|M|$ & 450 & $[4, 299]$ & $[5, 60]$ & $[2, 50]$ & integer & integer \\ 
		3 & network & network, with $|J|=|M|^*$ & 350 & $[10, 299]$ & $[10, 299]$ & $[2, 35]$ & integer & integer \\ 
		4 & planar & planar, with $|J|=|M|^*$ & 600 & $[10, 300]$ & $[10, 300]$ & $[2, 35]$ & float & integer \\
		\bottomrule
		\multicolumn{9}{l}{\scriptsize$^*$ The number of machine candidate locations are equal to the number of job storage locations.} \\
	\end{tabular}
	\label{tab:instances}
\end{table}

For the sets $1$, $2$ and $3$, the processing times and release dates of the jobs are represented by integer values, while for the set $4$, the release dates are represented by float numbers. For this particular set, our experiments were conducted by rounding down the release date values to consider two decimal places.

\subsection{Computational experiments}
\label{sec:experiments}

In order to assess the quality of the proposed formulation and algorithms, three sets of experiments have been performed. In the first set, we evaluate the proposed AF formulation by solving its linear relaxation and by comparing the obtained results with the ones obtained by two formulations from the related literature. In the second and third sets, we evaluate the heuristic methods described in Section \ref{sec:ils} and the exact framework algorithm presented in Section \ref{sec:exactframework}, respectively.

\subsubsection{Evaluating the AF formulation}
\label{sec:AFeval}

As stated before, the proposed AF formulation is characterized by having a pseudo-polynomial number of variables and constraints, demanding a high memory usage for modeling large scale problems. This may be a drawback when compared to the formulations by \citet{Hessler2017} and \citet{Wang2020}, whose number of variables and constraints are smaller, i.e., $O(n^2 m)$ and $O(n^2 m)$, for the first, and $O(n^2 m)$ and $O(n\max\{n,m\})$, for the second. However, the linear relaxation of the AF formulation is much tighter than those of its counterparts. This can be verified in Table \ref{tab:linrelax}, where it is shown for each formulation the average percentage gaps between the lower bounds and the optimal values (obtained by our framework algorithm, as reported in Section \ref{sec:Frameworkeval}). The percentage gap is computed as $100(opt-LB_{LR})/opt$, where $opt$ refers to the optimal value, and $LB_{LR}$ refers to the linear relaxation lower bound. 

Table \ref{tab:linrelax} reports the results for a set of 372 instances, from sets 1 (\texttt{small}) and 2 (\texttt{large}), with $n \leq 150$. Each row shows the average results for each group of instances defined according to the number of jobs $n$ (column \emph{\#jobs}). Columns \emph{\#inst} and \emph{opt.value} give the number of instances and the average optimal values associated to the group. 
Columns \emph{HD2017}, \emph{W2019}, and AF-CG under the label \emph{gap(\%)} report the mentioned gaps obtained by the linear relaxations of formulations by \citet{Hessler2017}, \citet{Wang2020}, and by the AF formulation \eqref{FO:AF}-\eqref{constr8:AF} solved by the CG procedure presented in Section \ref{sec:CG}, respectively, while those below the label \emph{time (s)} report the execution time.

\begin{table}[htbp]
	\centering
	\footnotesize
	\caption{Comparison of linear relaxations -- percentage gaps to optimal solutions}
	\scalebox{0.95}
	{
		\begin{tabular}{lrrHrrrrrrrr}
			\toprule
			\multirow{2}{*}{set} & \multirow{2}{*}{\#jobs} & \multirow{2}{*}{\# inst.} &  & opt. & \multicolumn{3}{r}{{gap (\%)}} & & \multicolumn{3}{r}{{time (s)}}\\
			\cline{6-8}\cline{10-12}
			&  &  & {iLB} & {value} & {HD2017} & {W2019} & AF-CG & & {HD2017} & {W2019} & AF-CG \\
			\midrule
			small & 1 $\leq n \leq$ 5 & 10    & 12.20 & 14.70 & 3.47  & 2.48  & \textbf{0.00} &       & 0.11  & $<$0.01  & 0.11 \\
			& 6 $\leq n \leq$ 10 & 13    & 19.31 & 21.38 & 29.68 & 29.34 & 0.74  &       & 0.11  & 0.02  & 0.17 \\
			& 11 $\leq n \leq$ 15 & 11    & 23.45 & 24.00 & 38.46 & 38.34 & 2.50  &       & 0.15  & 0.04  & 0.24 \\
			& 16 $\leq n \leq$ 20 & 6     & 29.17 & 29.33 & 44.51 & 44.22 & 1.39  &       & 0.20  & 0.12  & 0.18 \\
			& 21 $\leq n \leq$ 25 & 5     & 30.00 & 30.00 & 57.23 & 56.98 & \textbf{0.00} &       & 0.23  & 0.22  & 0.10 \\
			& 26 $\leq n \leq$ 30 & 5     & 60.20 & 60.20 & 74.62 & 74.42 & \textbf{0.00} &       & 0.31  & 0.43  & 0.10 \\
			\midrule
			large & 1 $\leq n \leq$ 10 & 4     & 23.50 & 27.75 & 18.10 & 18.10 & 3.57  &       & 0.11  & 0.02  & 0.12 \\
			& 11 $\leq n \leq$ 20 & 11    & 27.64 & 31.45 & 18.55 & 18.38 & 1.85  &       & 0.26  & 0.16  & 0.20 \\
			& 21 $\leq n \leq$ 30 & 23    & 28.96 & 31.22 & 23.32 & 23.32 & 1.22  &       & 0.81  & 0.64  & 0.21 \\
			& 31 $\leq n \leq$ 40 & 29    & 40.69 & 42.97 & 26.14 & 26.13 & 0.70  &       & 3.19  & 2.29  & 0.18 \\
			& 41 $\leq n \leq$ 50 & 35    & 46.63 & 47.09 & 42.65 & 42.64 & 0.34  &       & 3.31  & 4.70  & 0.34 \\
			& 51 $\leq n \leq$ 60 & 32    & 63.81 & 64.25 & 50.88 & 50.86 & \textbf{0.00} &       & 9.54  & 3.51  & 0.38 \\
			& 61 $\leq n \leq$ 70 & 34    & 74.47 & 74.59 & 53.16 & 53.15 & 0.22  &       & 13.80 & 5.34  & 0.18 \\
			& 71 $\leq n \leq$ 80 & 26    & 74.27 & 74.35 & 55.68 & 55.68 & 0.34  &       & 20.99 & 9.51  & 0.16 \\
			& 81 $\leq n \leq$ 90 & 23    & 81.96 & 82.00 & 59.27 & 59.25 & \textbf{0.00} &       & 38.37 & 15.49 & 0.75 \\
			& 91 $\leq n \leq$ 100 & 32    & 106.50 & 106.53 & 64.21 & 64.20 & 0.11  &       & 36.50 & 18.78 & 0.16 \\
			& 101 $\leq n \leq$ 110 & 11    & 63.73 & 63.73 & 60.75 & 60.75 & \textbf{0.00} &       & 85.83 & 29.44 & 0.14 \\
			& 111 $\leq n \leq$ 120 & 13    & 86.15 & 86.15 & 65.03 & 65.03 & \textbf{0.00} &       & 92.09 & 35.14 & 0.09 \\
			& 121 $\leq n \leq$ 130 & 19    & 77.26 & 77.26 & 65.70 & 65.70 & \textbf{0.00} &       & 150.94 & 51.69 & 0.12 \\
			& 131 $\leq n \leq$ 140 & 17    & 90.53 & 90.53 & 69.55 & 69.55 & \textbf{0.00} &       & 165.91 & 61.86 & 0.15 \\
			& 141 $\leq n \leq$ 150 & 13    & 96.00 & 96.00 & 70.88 & 70.88 & \textbf{0.00} &       & 164.47 & 66.27 & 0.14 \\
			\bottomrule
		\end{tabular}%
	}
	\label{tab:linrelax}%
\end{table}%

From Table \ref{tab:linrelax}, we observe that the lower bounds obtained by solving the linear relaxation of the proposed AF formulation are usually very close or equal to the optimal values, while the ones obtained by the other two formulations are far from them. This happens because \citet{Hessler2017} and \citet{Wang2020} use \emph{big-M} values to formulate the constraints associated to the jobs completion times. As the linear relaxation lower bounds of such formulations are very weak, solving them by means of branch-and-bound methods may require several branches and, consequently, high memory usage and large computing times. On the other hand, due to the good linear relaxation lower bounds, the AF formulation can be solved after a few branches and within shorter computing times. 
Although this experiment is limited to instances with up to $150$ jobs, these results suggest that AF formulation performs better than those from the current literature.

\subsubsection{Evaluating the heuristic methods}
\label{sec:heuristicsEval}

In this section, we assess the quality of the heuristic solutions obtained by the MIP-based heuristic methods \texttt{AF-CG} and \texttt{AF-subsetM}, and by the ILS algorithm presented in Section \ref{sec:heuristics}.

In order to evaluate the \texttt{AF-CG} method, we consider the columns obtained by the CG procedure when initialized with the solution generated by the deterministic procedure adopted in the constructive phase of our ILS algorithm. To create the restricted formulation used in the \texttt{AF-subsetM} method, we restrict the number of candidate locations to $\min\{|M|,\lfloor 1.5p\rfloor\}$ and select them using the same idea adopted in the deterministic constructive procedure of our ILS. Regarding the ILS algorithm, since it contains random components, it is evaluated based on the best and average results of 10 runs for each instance. In our experiments, the following parameter values were adopted: $n_\text{iter} = 10$, $n_\text{ils} = 100$, for the ILS heuristic, and a time limit of 120 seconds for solving the MIP-based heuristics. 

In Table \ref{tab:heuristics}, we report the summary of the obtained results and compare them with the best ones presented in the literature. Each row of the table shows the average percentage gap for each instance set. The percentage gap has been computed as $100(UB-LB)/LB$, where $UB$ refers to the solution value obtained by the heuristic procedure, and $LB$ refers to the best known lower bound. The values reported in columns \emph{HD2017$_\text{best}$} and \emph{W2019$_\text{best}$} refer to the best results obtained by \citet{Hessler2017} (over ten heuristics) and \citet{Wang2020} (over three heuristics), respectively. For the first, the best upper bounds are available online at \url{https://drive.google.com/open?id=0B_-MYN0r5mjqOEdGSEhoLW9Ubk0}. For the latter, the values have been collected from their article. The last four columns report the results obtained by the heuristics proposed in Section \ref{sec:heuristics}. Columns \texttt{AF-CG} and \texttt{AF-subsetM} refer to the heuristic results obtained by solving the MIP-based heuristics, while the columns labeled by \emph{ILS$_\text{avg}$} and \emph{ILS$_\text{best}$} refer to the average and best results obtained over the ten executions of the proposed ILS algorithm, respectively. In Table \ref{tab:proposedheur}, we report only the results for the proposed heuristics, including information regarding the computational times in seconds (labeled by columns \emph{time(s)}) and the number of optimal solutions found (labeled by columns \emph{\#opt.}). 

\begin{table}[h]
	\centering
	\footnotesize
	\caption{Heuristic results - percentage gaps}
	\scalebox{0.95}
	{
		\begin{tabular}{lrrHHHHHHHHHHHrrHHrrrrr}
			\toprule
			\multirow{2}{*}{instance set} & &                           & & & & & & & & & & & & &                          & & & & \multicolumn{4}{r}{this work} \\
			\cline{20-23}
			& & \up{HD2017$_\text{best}$} & & & & & & & & & & & & & \up{W2019$_\text{best}$} & & & & \texttt{AF-CG} & \texttt{AF-subsetM} & ILS$_\text{avg}$ & ILS$_\text{best}$ \\
			\cmidrule{1-1} \cmidrule{3-6} \cmidrule{9-11} \cmidrule{12-14} \cmidrule{16-18} \cmidrule{20-23}
			set 1 -- \texttt{small} & & 2.60\phantom{$^*$} & aaa & aaa & aaa &  & aaa & aaa & aaa &  & aaa & bbb & ccc &  & 1.93\phantom{$^*$} & eee & aaa &  & 1.38 & 0.42 & 0.39 & 0.00 \\ 
			set 2 -- \texttt{large} & & 10.70$^*$ & aaa & aaa & aaa &  & aaa & aaa & aaa &  & aaa & aaa & aaa &  & 4.10$^*$ & aaa & aaa &  & 0.91 & 0.18 & 0.39 & 0.20 \\ 
			set 3 -- \texttt{network} & & 7.54\phantom{$^*$} & aaa & aaa & aaa &  & aaa & aaa & aaa &  & aaa & aaa & aaa &  & 14.40$^*$ & aaa & aaa &  & 1.35 & 2.25 & 0.33 & 0.24 \\ 
			set 4 -- \texttt{planar} & & 6.92\phantom{$^*$} & aaa & aaa & aaa &  & aaa & aaa & aaa &  & aaa & aaa & aaa &  & 10.30$^*$ & aaa & aaa &  & 1.84 & 3.54 & 0.83 & 0.66 \\
			\bottomrule
			\multicolumn{23}{l}{$^*$ Values reported by \citet{Hessler2017} and \citet{Wang2020}.}\\
		\end{tabular}
	}
	\label{tab:heuristics}
\end{table}

From Table \ref{tab:heuristics}, we verify that the average results obtained by the proposed heuristics are always better than the best results from the related literature. Regarding the MIP-based heuristics, the small gaps are justified by the fact that the formulations are constructed from a good quality initial solution that allows to significantly reduce the number of variables and constraints, making the formulation easier to be solved, but still conserving a promising region of the solution space. This reduction is a contrast to the strategy adopted by \citet{Wang2020} in two heuristic procedures that consist of adding constraints to the formulation to restrict the search space. However, this strategy increases the size of the formulation and reduces, even more, the capacity to solve large-size instances. Regarding the ILS, the high quality results can be justified by the use of an efficient auxiliary data structure that allows performing local searches in a short computational time. On the contrary, the heuristics from the literature do not make use of auxiliary data structures, and either do not perform a local search (as those proposed by \citealt{Wang2020}) or do it over a very limited search space.

From Table \ref{tab:proposedheur}, we observe that the average execution times of the ILS algorithm for each set of instances is less than four seconds and that the ILS method was able to find optimal solutions for all instances from Set 1, and for more than 90\% of instances from the other sets.
Naturally, the MIP-based heuristics required more computational times and were able to find a smaller but reasonable number of optimal solutions when compared with the ILS. In particular, \texttt{AF-subsetM} performed better than \texttt{AF-CG} when solving instances from Sets 1 and 2, while the opposite occurred when solving Sets 3 and 4. It suggests that the deterministic procedure adopted to select the (candidate) machine locations does not perform well on instances of types network and planar (because they are characterized by having at least one release date equal to zero for each candidate locations), and that the variables obtained by the CG procedure can somehow attenuate this weakness. By grouping the instances according to the ratio $N/p$, we verify that the worst results occur when $N/p$ is less than five (see Table \ref{tab:pivot} in Appendix). As this holds for the four proposed heuristics, it suggests that the hardest instances are those with $N/p \leq 5$.

\begin{table}[htbp]
	\centering
	\footnotesize
	\caption{Results for the proposed heuristics}
	\setlength{\tabcolsep}{1mm}
	\scalebox{0.955}
	{
		\begin{tabular}{lrrrrrrrrrrrrrrr}
			\toprule
			instance &  & \multicolumn{ 3}{r}{\texttt{AF-CG}} &  & \multicolumn{ 3}{r}{\texttt{AF-subsetM}} &  & \multicolumn{ 3}{r}{ILS$_\text{avg}$} &  & \multicolumn{ 2}{r}{ILS$_\text{best}$} \\
			\cline{3-5} \cline{7-9} \cline{11-13} \cline{15-16}
			set &  & gap (\%) & time (s) & \#opt. &  & gap (\%) & time (s) & \#opt. &  & gap (\%) & time (s) & \#opt. &  & gap (\%) & \#opt. \\
			\cmidrule{1-1} \cmidrule{3-5} \cmidrule{7-9} \cmidrule{11-13} \cmidrule{15-16}
			\texttt{small} &  & 1.38 & 0.01 & 41 &  & 0.42 & 0.01 & 49 &  & 0.39 & 0.01 & 47.2 &  & 0.00 & 50 \\ 
			\texttt{large} &  & 0.91 & 0.99 & 351 &  & 0.18 & 13.79 & 410 &  & 0.39 & 0.82 & 404.8 &  & 0.20 & 426 \\ 
			\texttt{network} &  & 1.35 & 23.27 & 239 &  & 2.25 & 27.49 & 266 &  & 0.33 & 3.04 & 315.2 &  & 0.24 & 324 \\ 
			\texttt{planar} &  & 1.84 & 64.25 & 350 &  & 3.54 & 74.77 & 363 &  & 0.83 & 3.27 & 548.5 &  & 0.66 & 557 \\ \bottomrule
		\end{tabular}
	}
	\label{tab:proposedheur}
\end{table}

\subsubsection{Evaluating the exact framework}
\label{sec:Frameworkeval}

Finally, the results obtained by the exact framework (Algorithm \ref{algo:exact_scheloc}) are presented in Table \ref{tab:exact}. A time limit of 300 seconds was set in Gurobi for solving \texttt{AF-CG}, \texttt{AF-subsetM}, and the full AF formulation (\texttt{AF-full}), while the other parameters were kept the same as mentioned in the previous sections. The CG procedure is initialized with the columns associated with the ILS solution, and the subset of candidate locations $\overline{M}$ considered in the \texttt{AF-subsetM} method was defined as $\overline{M}=M''\cup R$, where $M''$ contains the $p$ machine locations chosen in the current best solution, and $R$ contains $\min\{\lfloor0.5p\rfloor, m-p\}$ locations randomly chosen from $M \setminus M''$. 

The results reported in Table \ref{tab:exact} are aggregated in groups of instances, classified according to the set they belong and to the ratio $N/p$, as shown in the first two columns. The number of instances in each group is given by the column ``\emph{\#inst.}''. Since all instances have been solved to optimality, the four columns below the label ``\emph{\#opt.}'' report the number of instances whose optimal solutions have been found after running the procedure indicated in the column header, i.e., the \texttt{ILS}, the \texttt{AF-CG}, the \texttt{AF-subsetM}, and the \texttt{AF-full}. The order in which they are presented in the table, from left to right, is the same as they are executed by the framework. The last column, ``\emph{avg. time (s).}'', reports the average running time, in seconds, for each group of instances.
\begin{table}[!h]
	\centering
	\caption{Results for the exact framework}
	\setlength{\tabcolsep}{3.5mm}
	\scalebox{0.77}
	{
		\begin{tabular}{lrrlrrrrr}
			\toprule
			\multicolumn{3}{r}{instance group} & & \multicolumn{4}{r}{\#opt.} &  \\
			\cmidrule{1-3} \cmidrule{5-8}
			set & $N/p$ & \#inst. &  & \texttt{ILS} & \texttt{AF-CG} & \texttt{AF-subsetM} & \texttt{AF-full} & \up{avg. time (s)} \\
			\cmidrule{1-3} \cmidrule{5-9}
			\texttt{small} & ( \phantom{x}0, \phantom{x}1 ] & 5 &  & 0 & 5 & 0 & 0 & 0.01 \\ 
			& ( \phantom{x}1, \phantom{x}2 ] & 11 &  & 2 & 7 & 2 & 0 & 0.03 \\ 
			& ( \phantom{x}2, \phantom{x}3 ] & 4 &  & 2 & 1 & 1 & 0 & 0.03 \\ 
			& ( \phantom{x}3, \phantom{x}4 ] & 8 &  & 6 & 0 & 1 & 1 & 0.02 \\ 
			& ( \phantom{x}4, \phantom{x}5 ] & 6 &  & 5 & 0 & 0 & 1 & 0.01 \\ 
			& ( \phantom{x}5, 10 ] & 12 &  & 12 & 0 & 0 & 0 & 0.00 \\ 
			& ( 10, 15 ] & 4 &  & 4 & 0 & 0 & 0 & 0.00 \\
			\midrule
			\texttt{large} & ( \phantom{x}0, \phantom{x}1 ] & 3 &  & 0 & 3 & 0 & 0 & 0.26 \\ 
			& ( \phantom{x}1, \phantom{x}2 ] & 35 &  & 0 & 26 & 4 & 5 & 0.95 \\ 
			& ( \phantom{x}2, \phantom{x}3 ] & 41 &  & 6 & 9 & 20 & 6 & 2.33 \\ 
			& ( \phantom{x}3, \phantom{x}4 ] & 65 &  & 49 & 5 & 10 & 1 & 3.54 \\ 
			& ( \phantom{x}4, \phantom{x}5 ] & 50 &  & 49 & 0 & 1 & 0 & 0.59 \\ 
			& ( \phantom{x}5, 10 ] & 142 &  & 142 & 0 & 0 & 0 & 0.75 \\ 
			& ( 10, 15 ] & 55 &  & 55 & 0 & 0 & 0 & 0.32 \\ 
			& ( 15, 20 ] & 42 &  & 42 & 0 & 0 & 0 & 0.24 \\ 
			& ( 20, 25 ] & 7 &  & 7 & 0 & 0 & 0 & 0.14 \\ 
			& ( 25, 50 ] & 10 &  & 10 & 0 & 0 & 0 & 0.02 \\
			\midrule
			\texttt{network} & ( \phantom{x}2, \phantom{x}3 ] & 18 &  & 4 & 2 & 4 & 8 & 1.91 \\ 
			& ( \phantom{x}3, \phantom{x}4 ] & 38 &  & 24 & 7 & 5 & 2 & 10.37 \\ 
			& ( \phantom{x}4, \phantom{x}5 ] & 37 &  & 25 & 4 & 6 & 2 & 39.05 \\ 
			& ( \phantom{x}5, 10 ] & 174 &  & 164 & 3 & 7 & 0 & 17.96 \\ 
			& ( 10, 15 ] & 64 &  & 64 & 0 & 0 & 0 & 2.73 \\ 
			& ( 15, 20 ] & 19 &  & 19 & 0 & 0 & 0 & 1.82 \\
			\midrule
			\texttt{planar} & ( \phantom{x}1, \phantom{x}2 ] & 1 &  & 0 & 0 & 0 & 1 & 0.38 \\ 
			& ( \phantom{x}2, \phantom{x}3 ] & 22 &  & 10 & 0 & 0 & 12 & 19.22 \\ 
			& ( \phantom{x}3, \phantom{x}4 ] & 38 &  & 14 & 3 & 9 & 12 & 116.12 \\ 
			& ( \phantom{x}4, \phantom{x}5 ] & 52 &  & 36 & 1 & 10 & 5 & 114.51 \\ 
			& ( \phantom{x}5, 10 ] & 300 &  & 295 & 0 & 5 & 0 & 7.58 \\ 
			& ( 10, 15 ] & 155 &  & 155 & 0 & 0 & 0 & 2.28 \\ 
			& ( 15, 20 ] & 29 &  & 29 & 0 & 0 & 0 & 1.46 \\ 
			& ( 20, 25 ] & 3 &  & 3 & 0 & 0 & 0 & 0.00 \\
			\midrule
			\multicolumn{2}{l}{\textbf{sum/avg.}} & {\textbf{1450}} & & {\textbf{1233}} & {\textbf{76}} & {\textbf{85}} & {\textbf{56}} & {\textbf{2.23}} \\ \bottomrule 
		\end{tabular}
	}
	\label{tab:exact}
\end{table}

From Table \ref{tab:exact}, we observe that 85\% of the instances were solved to optimality by the ILS procedure, which has found solutions with the same cost as the lower bound computed by Equation \ref{eq:initLB}. From the remaining 217 instances, \texttt{AF-GC} solved 76 to optimality. Then, \texttt{AF-subsetM} solved 85 out of 141, and \texttt{AF-full} solved the remaining 56 ones. As can be noticed, executing less-complex procedures before solving the full AF formulation makes it possible to save CPU time and to solve large size instances that would be intractable by exact models due to the high demand for memory. Moreover, for the few cases that the full AF model was solved, it started with very good lower and upper bounds (obtained by the previously executed procedures), saving CPU time and memory resources, especially when solving instances with $|N|=|M|$ and with release dates assuming float values, i.e., from sets 3 (\texttt{network}) and 4 (\texttt{planar}). 

\subsection{Experiments on new challenging instances}
\label{sec:newinst}

Since all benchmark instances from the related literature have been solved to optimality, we created a new set of instances to further evaluate the exact framework algorithm. These new instances have characteristics similar to those from the \texttt{planar} set proposed by \citet{Hessler2017}, but with $N \neq M$ and with larger processing times. We considered different combinations of $n$, $m$ and $p$, with $n \in \{100,200,300\}$, $m \in \{20,40,60,$ $80,100,120,140,160,180,200,220,240,260,280\}$, and $p \in \{2,3,5,7,10,15,20,30,40,$ $50,75,100\}$, such that $p < m < n$. The coordinates for the jobs ($x_j$, $y_j$) and for the machine candidate locations ($x_k, y_k$) have been randomly generated within a square $n \times n$, and the job processing times have been randomly generated between $\lfloor n/10 \rfloor$ and $\lfloor n/2 \rfloor$, both considering an uniform distribution. Then, based on the coordinates of job $j$ and machine candidate location $k$, we computed the release dates $r_{jk}$, for every $j \in J$ and $k \in M$, as $r_{jk} = \big\lfloor \sqrt{(x_j - x_k)^2+(y_j-y_k)^2} \big\rfloor$. In total, 283 instances were generated, 35 of which with $n=100$, 94 with $n=200$, and 154 with $n=300$. The new generated instances are available at \url{https://github.com/raphaelhk/ScheLoc-instances.git}.

The results obtained by the exact framework for the new set of instances are reported in Table \ref{tab:newinst}, which is organized similarly to Table \ref{tab:exact}, with the addition of one column to report the percentage gaps. As can be observed, only $34\%$ ($96$ out of $283$) of the instances were solved to optimality, suggesting that they are indeed more challenging than those proposed by \citet{Hessler2017}. This fact may be justified by the fact that in the new instances the jobs have larger and more distinct processing times, which lead to large makespans ($C_\text{max}$) and somehow weakens the quality of the lower bounds obtained by Equation \eqref{eq:initLB} and by the linear relaxation of the mathematical models. During the execution of the framework, if the computed lower bounds are smaller than the optimal value, it is not possible to prove the optimality without solving the full AF formulation. 

The arguments mentioned above are supported by the small number of instances solved to proven optimality before the execution of \texttt{AF-full}. Also, since the size of the AF formulation depends on the value of $C_{\max}$, which is expected to be large on these new instances, a considerable amount of CPU time and memory could be required to prove the optimality. This fact is evidenced by the small number of instances solved to proven optimality after executing \texttt{AF-full}. Moreover, none of the models associated with the instances with $n=300$ could be solved due to memory limitations. Despite this, the proposed framework algorithm was able to provide very good upper and lower bounds as can be verified by the small percentage gaps, and, similarly to what was observed in Section \ref{sec:heuristicsEval}, we verify that the hardest instances are those with small values of $N/p$. 
\begin{table}[!h]
	\centering
	\caption{Results for the new set of instances}
	\setlength{\tabcolsep}{3.5mm}
	\scalebox{0.82}
	{
		\begin{tabular}{lrrlrrrrrr}
			\toprule
			\multicolumn{3}{r}{instance group} & & \multicolumn{4}{r}{\#opt.} &  \\
			\cmidrule{1-3} \cmidrule{5-8}
			$N$ & $N/p$ & \#inst. &  & \texttt{ILS} & \texttt{AF-CG} & \texttt{AF-subsetM} & \texttt{AF-full} & \up{avg. time (s)} & \up{gap (\%)} \\ \cmidrule{1-3} \cmidrule{5-10}
			100 & ( \phantom{x}0, \phantom{x}2 ] & 3 &  & 0 & 0 & 2 & 1 & 34.17 & 0.00 \\ 
			& ( \phantom{x}2, \phantom{x}3 ] & 2 &  & 0 & 0 & 1 & 1 & 50.35 & 0.00 \\ 
			& ( \phantom{x}3, \phantom{x}4 ] & 3 &  & 0 & 0 & 1 & 2 & 156.77 & 0.00 \\ 
			& ( \phantom{x}4, \phantom{x}5 ] & 3 &  & 0 & 0 & 0 & 2 & 525.28 & 0.23 \\ 
			& ( \phantom{x}5, 10 ] & 8 &  & 1 & 1 & 1 & 2 & 322.50 & 0.18 \\ 
			& ( 10, 15 ] & 4 &  & 2 & 0 & 0 & 0 & 221.01 & 0.12 \\ 
			& ( 15, 20 ] & 4 &  & 2 & 0 & 0 & 0 & 295.29 & 0.09 \\ 
			& ( 20, 50 ] & 8 &  & 8 & 0 & 0 & 0 & 0.10 & 0.00 \\ \midrule
			200 & ( \phantom{x}0, \phantom{x}2 ] & 4 &  & 0 & 0 & 1 & 3 & 657.19 & 0.00 \\ 
			& ( \phantom{x}2, \phantom{x}3 ] & 6 &  & 0 & 0 & 0 & 0 & 950.23 & 3.47 \\ 
			& ( \phantom{x}3, \phantom{x}4 ] & 7 &  & 0 & 0 & 0 & 0 & 1,052.31 & 1.80 \\ 
			& ( \phantom{x}4, \phantom{x}5 ] & 7 &  & 0 & 0 & 0 & 0 & 909.72 & 0.84 \\ 
			& ( \phantom{x}5, 10 ] & 16 &  & 0 & 0 & 0 & 0 & 901.52 & 0.43 \\ 
			& ( 10, 15 ] & 9 &  & 1 & 0 & 0 & 0 & 692.77 & 0.16 \\ 
			& ( 15, 20 ] & 9 &  & 0 & 0 & 0 & 0 & 643.72 & 0.11 \\ 
			& ( 20, 50 ] & 36 &  & 25 & 0 & 0 & 0 & 197.64 & 0.02 \\ \midrule
			300 & ( \phantom{x}0, \phantom{x}3 ] & 9 &  & 0 & 0 & 0 & 0 & 384.80 & 2.56 \\ 
			& ( \phantom{x}3, \phantom{x}4 ] & 11 &  & 0 & 0 & 0 & 0 & 265.44 & 1.13 \\ 
			& ( \phantom{x}4, 10 ] & 37 &  & 0 & 0 & 0 & 0 & 111.37 & 0.38 \\ 
			& ( 10, 15 ] & 13 &  & 0 & 0 & 0 & 0 & 41.62 & 0.15 \\ 
			& ( 15, 20 ] & 14 &  & 0 & 0 & 0 & 0 & 28.58 & 0.07 \\ 
			& ( 20, 50 ] & 70 &  & 39 & 0 & 0 & 0 & 7.65 & 0.01 \\ \midrule
			\multicolumn{2}{l}{\textbf{sum/avg.}} & \textbf{283} & & \textbf{78} & \textbf{1} & \textbf{6} & \textbf{11} & \textbf{384.09} & \textbf{0.37} \\
			\bottomrule 
		\end{tabular}
	}
	\label{tab:newinst}
\end{table}

\section{Conclusions}
\label{sec:conclusions}

In this paper, we have presented and evaluated some procedures for solving the discrete parallel machine makespan scheduling-location problem. For instance, a new arc-flow based mathematical formulation, a column generation, and three heuristic methods have been proposed to solve the problem. 
It was shown through computational experiments that the arc-flow formulation is stronger than the existing ones from the current literature and that the proposed heuristics are capable of generating high-quality solutions. In particular, the ILS algorithm enhanced with auxiliary data structures, that allow us to perform move evaluations in constant time, was able to find optimal solutions for 93\% of the benchmark instances from the literature in a short computational time. 

By embedding such procedures into a framework algorithm, it was capable of obtaining optimal solutions for all existing benchmark instances from the related literature, most of them for the first time. The framework was also tested on a new set of $283$ large and challenging instances, among which $187$ of them remain unsolved to proven optimality.

Future researches can be carried out to develop tailored exact methods, such as branch-and-cut and branch-and-price algorithms, to deal with the challenging instances or even on the adaptation of the proposed procedures to solve variants of the problem, such as the DPMM ScheLoc problem with unrelated machines.


\bibliographystyle{mmsbib}
\bibliography{ref}


\newpage
\appendix

\section{Heuristic results classified by the ratio $N/p$}
\label{sec:pivot}


\begin{table}[htbp]
	\centering
	\caption{Heuristic results classified by the ratio $N/p$}
	\scalebox{0.68}
	{
		\begin{tabular}{rrrrrrrrrrrrrrrrrr}
			\hline
			\multicolumn{3}{r}{instance} & & \multicolumn{3}{r}{AF-CG} & & \multicolumn{3}{r}{AF-subsetM} & & \multicolumn{3}{r}{ILS-avg} & & \multicolumn{2}{r}{ILS-best} \\ \cline{1-3} \cline{5-7} \cline{9-11} \cline{13-15} \cline{17-18}
			set & $N/p$ & \#inst. & & gap & time (s) & \#opt. & & gap & time (s) & \#opt. & & gap & time (s) & \#opt. & & gap & \#opt. \\ \cline{1-3} \cline{5-7} \cline{9-11} \cline{13-15} \cline{17-18}
			Set 1 & (0, 1] & 5 &  & 0.00 & 0.00 & 5 &  & 0.00 & 0.00 & 5 &  & 0.00 & 0.00 & 5 &  & 0.00 & 5 \\ 
			& (1, 2] & 11 &  & 0.00 & 0.01 & 11 &  & 0.00 & 0.00 & 11 &  & 0.87 & 0.02 & 9.8 &  & 0.00 & 11 \\ 
			& (2, 3] & 4 &  & {6.70} & 0.01 & 2 &  & 0.00 & 0.02 & 4 &  & 0.85 & 0.02 & 3.6 &  & 0.00 & 4 \\ 
			& (3, 4] & 8 &  & {4.42} & 0.01 & 3 &  & 0.00 & 0.01 & 8 &  & 0.56 & 0.01 & 7.2 &  & 0.00 & 8 \\ 
			& (4, 5] & 6 &  & 0.00 & 0.01 & 6 &  & {3.47} & 0.00 & 5 &  & 0.00 & 0.00 & 6 &  & 0.00 & 6 \\ 
			& (5, 10] & 12 &  & 0.56 & 0.01 & 10 &  & 0.00 & 0.01 & 12 &  & 0.17 & 0.00 & 11.6 &  & 0.00 & 12 \\ 
			& (10, 15] & 4 &  & 0.00 & 0.00 & 4 &  & 0.00 & 0.01 & 4 &  & 0.00 & 0.00 & 4 &  & 0.00 & 4 \\ \hline
			Set 2 & (0, 1] & 3 &  & {1.09} & 0.01 & 2 &  & {1.09} & 0.03 & 2 &  & {1.09} & 0.27 & 2 &  & {1.09} & 2 \\ 
			& (1, 2] & 35 &  & {2.33} & 0.05 & 27 &  & 0.00 & 0.18 & 35 &  & {1.22} & 0.92 & 26.1 &  & 0.67 & 30 \\ 
			& (2, 3] & 41 &  & {4.67} & 0.28 & 9 &  & 0.44 & 4.01 & 39 &  & {2.21} & 1.86 & 19.9 &  & {1.24} & 28 \\ 
			& (3, 4] & 65 &  & {1.09} & 0.69 & 48 &  & 0.07 & 13.13 & 63 &  & 0.53 & 1.12 & 53.4 &  & 0.18 & 61 \\ 
			& (4, 5] & 50 &  & 0.40 & 2.40 & 41 &  & 0.43 & 27.81 & 40 &  & 0.09 & 0.63 & 47.6 &  & 0.04 & 49 \\ 
			& (5, 10] & 142 &  & 0.27 & 1.73 & 115 &  & 0.22 & 23.23 & 120 &  & 0.00 & 0.84 & 141.8 &  & 0.00 & 142 \\ 
			& (10, 15] & 55 &  & 0.06 & 0.33 & 50 &  & 0.04 & 8.19 & 52 &  & 0.00 & 0.37 & 55 &  & 0.00 & 55 \\ 
			& (15, 20] & 42 &  & 0.00 & 0.02 & 42 &  & 0.00 & 0.76 & 42 &  & 0.00 & 0.29 & 42 &  & 0.00 & 42 \\ 
			& (20, 25] & 7 &  & 0.00 & 0.02 & 7 &  & 0.00 & 0.72 & 7 &  & 0.00 & 0.16 & 7 &  & 0.00 & 7 \\ 
			& (25, 50] & 10 &  & 0.00 & 0.02 & 10 &  & 0.00 & 0.30 & 10 &  & 0.00 & 0.02 & 10 &  & 0.00 & 10 \\ \hline
			Set 3 & (2, 3] & 18 &  & {10.46} & 0.43 & 5 &  & {10.28} & 1.72 & 8 &  & {1.75} & 1.32 & 11.3 &  & {1.44} & 12 \\ 
			& (3, 4] & 38 &  & {2.15} & 3.49 & 25 &  & {3.90} & 11.92 & 31 &  & {1.05} & 1.92 & 26.8 &  & 0.83 & 29 \\ 
			& (4, 5] & 37 &  & {1.71} & 17.30 & 21 &  & {6.25} & 20.10 & 31 &  & 0.91 & 5.56 & 25.8 &  & 0.67 & 28 \\ 
			& (5, 10] & 174 &  & 0.69 & 24.95 & 127 &  & {1.19} & 33.56 & 133 &  & 0.06 & 3.01 & 168.3 &  & 0.02 & 172 \\ 
			& (10, 15] & 64 &  & 0.24 & 38.84 & 45 &  & 0.22 & 33.93 & 47 &  & 0.00 & 3.11 & 64 &  & 0.00 & 64 \\ 
			& (15, 20] & 19 &  & 0.10 & 28.40 & 16 &  & 0.10 & 20.17 & 16 &  & 0.00 & 2.01 & 19 &  & 0.00 & 19 \\ \hline
			Set 4 & (1, 2] & 1 &  & {7.14} & 0.13 & 0 &  & 0.00 & 0.08 & 1 &  & 0.00 & 0.03 & 1 &  & 0.00 & 1 \\ 
			& (2, 3] & 22 &  & {12.61} & 3.96 & 3 &  & {26.49} & 6.74 & 6 &  & {9.10} & 2.78 & 10.2 &  & {7.57} & 11 \\ 
			& (3, 4] & 38 &  & {7.96} & 28.73 & 8 &  & {20.81} & 58.20 & 11 &  & {5.59} & 7.98 & 19.1 &  & {4.57} & 21 \\ 
			& (4, 5] & 52 &  & {4.79} & 81.26 & 19 &  & {8.33} & 70.06 & 19 &  & {1.30} & 5.23 & 38.7 &  & {1.00} & 40 \\ 
			& (5, 10] & 300 &  & 0.78 & 73.80 & 175 &  & 0.95 & 75.02 & 180 &  & 0.05 & 3.08 & 292.5 &  & 0.02 & 297 \\ 
			& (10, 15] & 155 &  & 0.19 & 62.15 & 118 &  & 0.18 & 86.89 & 119 &  & 0.00 & 2.30 & 155 &  & 0.00 & 155 \\ 
			& (15, 20] & 29 &  & 0.09 & 47.16 & 24 &  & 0.09 & 99.47 & 24 &  & 0.00 & 1.45 & 29 &  & 0.00 & 29 \\ 
			& (20, 25] & 3 &  & 0.00 & 2.03 & 3 &  & 0.00 & 0.43 & 3 &  & 0.00 & 0.00 & 3 &  & 0.00 & 3 \\ \hline
		\end{tabular}
	}
	\label{tab:pivot}
\end{table}
\end{document}